\documentclass{article}
\usepackage{iclr2027_conference,times}

\usepackage{amsmath,amsfonts,bm}

\def\eqref#1{equation~\ref{#1}}

\def\1{\bm{1}}

\DeclareMathAlphabet{\mathsfit}{\encodingdefault}{\sfdefault}{m}{sl}
\SetMathAlphabet{\mathsfit}{bold}{\encodingdefault}{\sfdefault}{bx}{n}

\usepackage{hyperref,url,booktabs,graphicx,amsmath,amssymb,xcolor,subcaption,tikz}
\usepackage{colortbl}
\usepackage{placeins}
\definecolor{oursbg}{HTML}{FBE3DA}
\definecolor{ablbg}{HTML}{FDF2EE}
\usetikzlibrary{arrows.meta,positioning,fit,calc}

\newif\ifarxiv\arxivtrue
\newcommand{\method}{\textsc{AbridgeKV}}
\iclrfinalcopy

\title{\method{}: Compressing Long Reasoning\\into the KV Cache of Learned Pause Tokens}
\author{Jack Cai$^{1}$\thanks{Correspondence to \texttt{zc5794@princeton.edu}.}\hspace{0.9em}Lijie Yang$^{1}$\hspace{0.6em}Vasilis Kontonis$^{2}$\hspace{0.6em}Dimitris Papailiopoulos$^{2}$\hspace{0.6em}Shivam Garg$^{2}$\\[0.6ex]
\mdseries $^{1}$Princeton University\qquad $^{2}$Microsoft Research}

\begin{document}
\maketitle
\lhead{Preprint}
\thispagestyle{fancy}

\begin{abstract}
Long chain-of-thought decoding is memory-bound: the KV cache of the generated tokens grows linearly with the test-time
compute being spent. Evicting cache entries does not fix this, because a model never trained on a cache with entries removed
generates longer, often up to the length limit. We instead train the model to compress its own output. \method{} compresses each completed block of generated text into the KV
entries of a few learned pause tokens. These are extra input tokens with trained embeddings that produce no output. Once the
pause tokens have encoded the block, \method{} frees the block's original cache; subsequent tokens attend to the pause
tokens' KV entries in its place. Three novel choices make this work: the pause tokens' KV entries take positions inside the
block they replace, the cache budget is allocated across layers according to the measured value of each layer's entries, and
compression operates on relatively large blocks. On 446 competition problems from AIME, HMMT, AMC, BRUMO, CMIMC and SMT with 32K-token generations,
keeping a quarter of the generated cache matches an identically trained full-KV model in accuracy and output length; the
strongest published evictor at a matched budget writes far longer, and the closest trained compressor, retrained on the same
data, loses 10 points. Served in vLLM, the cache is $3.5\times$ smaller per sequence and a saturated H100 finishes a problem $3.2\times$
faster than with full KV.
\end{abstract}

\section{Introduction}
\label{sec:intro}
Test-time scaling improves reasoning accuracy by generating longer chains of thought
(CoT)~\citep{openai2024o1,deepseek2025r1,muennighoff2025s1,snell2024scaling}. On competition mathematics, even small models
generate tens of thousands of tokens per response and frequently reach the maximum generation length. Decoding such long
CoTs is memory-bound: the KV cache grows linearly with the number of generated tokens, and its size, not the model weights,
bounds the batch size and hence throughput~\citep{kwon2023vllm}. Unlike long-prompt settings, this cache is produced by the
model's own decoding and grows with exactly the test-time compute being spent. We target the KV cache of the generated tokens.

\begin{figure}[t]
  \centering
  \includegraphics[width=\linewidth]{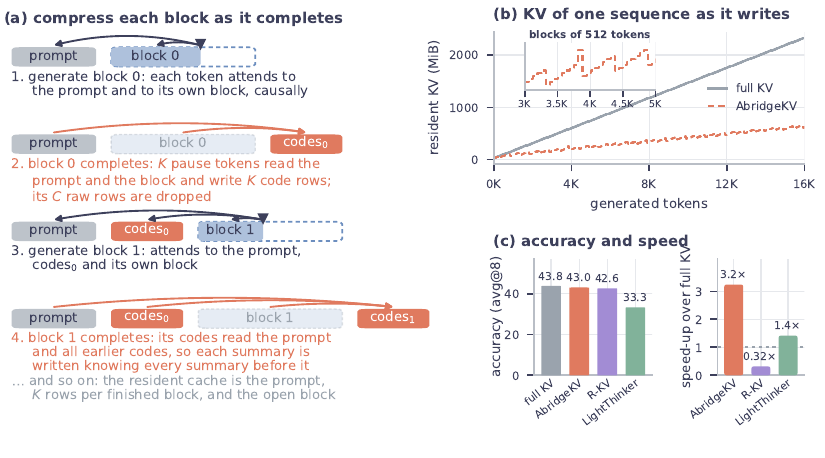}
  \caption{(a)~\method{} writes in blocks of $C$ tokens; when a block completes, $K$ pause tokens write its code rows (their KV entries,
  one per pause token per layer) and its raw rows are freed. (b)~KV of one sequence in vLLM: it rises through each block and falls at each compression. (c)~Accuracy (avg@8) on 446 competition problems at the 32K cap, and speed-up in time to finish a problem relative to full KV
  (dotted line) on a saturated H100.}
  \label{fig:teaser}
\end{figure}

KV eviction keeps a fixed number of cache entries and discards the rest, selected by recency or accumulated attention
score~\citep{zhang2023h2o,xiao2024streamingllm,li2024snapkv}; several recent methods specialize it to long
reasoning~\citep{cai2025rkv,song2025rpc,zhang2026lazyeviction,chang2026vase}. Two properties limit it. The model is never
trained on a cache with entries removed, and the selection is made from past attention, whereas a CoT reuses intermediate
results at unpredictable later points; an evicted entry leaves no representation behind. Empirically the failure appears as longer generations. When they keep a quarter of the cache, simple eviction policies (recency,
random, accumulated attention) let 38--58\% of samples run to the 32K limit, and R-KV, the strongest published method we
evaluate, generates 40--69\% more tokens than the uncompressed model at the two budgets we test (Figure~\ref{fig:lenflow}a), which offsets the per-token
memory saving. What to keep from a block of reasoning is better decided by a model trained for it: the compression should be
learned, and the model should be trained to continue from the compressed history.

Learned compression~\citep{mu2023gist,ge2024icae,zhang2024activationbeacon} was developed for fixed inputs, such as long
documents, compressed once before decoding. Compressing a CoT differs: the model must compress its own output online, dozens
of times per response, and condition on the compressed states thousands of tokens later, so each summary has to be written
without knowing which of its contents the later reasoning will need.

MEMENTO~\citep{kontonis2026memento}, which motivated this work, belongs to a line of work that trains models to compress
their own reasoning in text, including InftyThink~\citep{yan2025inftythink}, InftyThink+~\citep{yan2026inftythinkplus},
Accordion-Thinking~\citep{yang2026accordion} and The Markovian Thinker~\citep{aghajohari2025markovian}. MEMENTO splits the
model's reasoning into blocks, writes a short summary of each block, and continues from the summaries alone. Producing its
training data, however, requires a sophisticated annotation pipeline in which a frontier LLM scores candidate block
boundaries, writes the summaries, and grades and refines them. Can we avoid such a pipeline entirely and instead use a fixed
set of tokens as summaries, allowing their KV entries to learn from the next-token loss what information to preserve?

LightThinker~\citep{zhang2025lightthinker} and Breadcrumbs Reasoning~\citep{monea2025breadcrumbs} pursue this basic idea:
rather than producing explicit textual summaries, they replace generated text with the KV entries of learned memory tokens.
Their evaluations, however, involve shorter generations and less demanding long-reasoning settings. LightThinker is evaluated
with outputs of at most 10K tokens, while Breadcrumbs is evaluated on synthetic tasks with outputs of at most 1{,}000 tokens.
When we retrain LightThinker on our data, serve it in our engine, and evaluate it on long competition-math problems, its
accuracy is 10 points below the full-KV model. This drop is perhaps unsurprising: a small, fixed set of KV entries must
replace the information carried by an explicit textual summary. More surprisingly, we show that three design choices can
recover this loss and make pause-token compression effective at long generation lengths.

Our method, \method{}, compresses the generated KV cache in fixed blocks of 512 tokens (Figure~\ref{fig:teaser}a). When a
block is complete, a set of pause tokens attends to it and encodes a summary in their own KV entries; we call the KV entry of
one pause token at one layer a \emph{code row}. The block's original KV entries are then freed, and later tokens attend to
the code rows in their place. The model is trained end to end with the next-token loss, so later tokens teach the code rows
what information to preserve. Three design choices are central: code rows take positions inside the block they replace, the
cache budget is allocated across layers according to the measured value of each layer's entries, and compression operates
on relatively large blocks. Together, these choices make pause-token compression effective over long generations.

The resulting reduction is realized as physical memory savings in our paged serving engine. Because each layer has a fixed
budget, vLLM can maintain a separate page pool for each budget level: rows that a layer does not retain are never allocated,
and pages belonging to a freed block are returned to the pool for other requests.

\noindent\begin{minipage}{\linewidth}
Our contributions are as follows.
\begin{enumerate}\itemsep0pt
  \item We introduce \method{}, in which a reasoning model \textbf{compresses its own chain of thought into the KV cache of
  learned pause tokens} as it writes, trained end to end and served as physical memory in a paged engine (Section~\ref{sec:method}).
  \item We identify \textbf{three novel design choices}: in-block code positions (no prior learned summary uses them), a layer
  budget set by measured value (new to learned compression) and large blocks (the nearest prior method compresses every paragraph, about 50 tokens),
  and show that code rows carry their block's content many blocks later (Sections~\ref{sec:ablations}--\ref{sec:analysis}).
  \item At a 32K cap, \method{} \textbf{matches full KV with a $4\times$ smaller persistent cache and finishes a problem
  $3.2\times$ faster} in our tuned vLLM, where eviction writes far longer and LightThinker loses 10 points (Section~\ref{sec:results}).
\end{enumerate}
\end{minipage}
\ifarxiv
\paragraph{Future directions and implications.} \method{} suggests several promising directions beyond the setting studied
here. We highlight two that appear particularly compelling.

First, \method{} could work alongside the compaction mechanisms already used by long-running coding agents. Modern agents
typically compact their context only after it reaches roughly 250K to 1M tokens, rewriting accumulated file contents, tool
outputs, and reasoning into a textual summary before generation resumes. \method{} operates at a much finer scale: by
compressing every completed block, it reduces the rate at which each request consumes KV memory throughout the long interval
before the next compaction. At our 4:1 budget, for example, the generated-token KV grows at roughly one quarter of its
original rate. The two mechanisms therefore act at different scales and can naturally be combined. A more ambitious direction
is to ask whether recursive \method{}, compressing earlier code rows into new code rows, could eventually replace the
compaction step altogether and keep the cache bounded without rewriting the history as text.

Second, learned KV compression may be especially promising for modalities that are difficult to summarize faithfully in
text. Computer-use agents, for example, must reason over long sequences of screenshots and other visual observations whose
task-relevant content may not admit a natural textual summary. In such settings, the KV space may provide a more natural
compression medium, allowing the model to learn directly which visual information must be retained for future actions.
Extending \method{} to multimodal and computer-use agents is therefore an exciting direction for future work.
\fi

\section{Method}
\label{sec:method}
\begin{figure}[t]
  \centering
  \includegraphics[width=\linewidth]{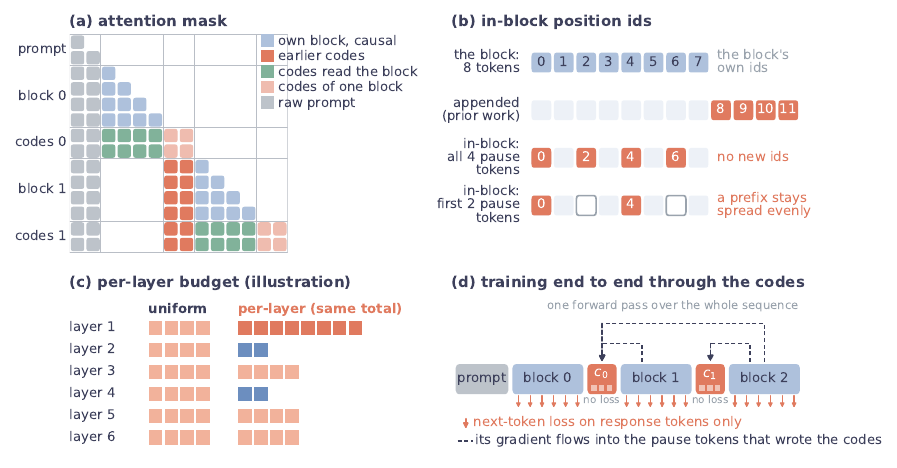}
  \caption{Method details. (a)~Attention mask for two blocks ($C{=}4$, $K{=}2$, 2-token prompt). (b)~Pause token $k$ (whose KV entries are code row $k$ at every layer) takes position
  $\mathrm{bitrev}(k)\cdot C/K_{\max}$ inside the block it replaces ($C{=}8$, $K_{\max}{=}4$), not after it as in prior work;
  any prefix of the pause tokens is an even grid. (c)~Per-layer quotas keep the uniform mean. (d)~Training is one forward pass under
  mask (a), with next-token loss on response tokens only; later blocks' loss reaches the pause tokens that wrote their codes.}
  \label{fig:layout}
\end{figure}

\subsection{Block layout and attention mask}
\paragraph{Layout.} The prompt ($P$ tokens) is kept raw and never compressed. The response is generated in blocks of $C{=}512$
tokens; block $b\in\{0,1,\dots\}$ has content $x_{b,0},\dots,x_{b,C-1}$ and $K_{\max}{=}256$ \emph{code slots}
$p_{b,0},\dots,p_{b,K_{\max}-1}$, pause tokens with learned input embeddings. When block $b$ completes, the $K_{\max}$ pause tokens
are run through the model in one forward pass; their KV rows at every layer are the block's \emph{codes}. The block's content KV is
then dropped. Nothing else is added to the model: no adapters, no extra layers, no change to the rotary embedding.

\paragraph{Visibility.} A content query $x_{b,t}$ attends to the prompt, to the codes of every earlier block $b'<b$, and to its own
block causally, $x_{b,<t}$. A code query $p_{b,k}$ attends to the prompt, to the codes of earlier blocks, to all of its block
$x_{b,0..C-1}$, and to the other codes of its block. Content of block $b$ is never visible to any query outside block $b$
(Figure~\ref{fig:layout}a). This mask is what the trainer enforces on the full sequence and what the runtime enforces by eviction.

\paragraph{Positions.} Content keeps its uncompressed numbering, and each code row sits \emph{inside} the block it summarizes.
The simplest such layout spaces the $K_{\max}$ pause tokens evenly over the block, pause token $k$ at offset $k\,C/K_{\max}$.
Even spacing alone already keeps every position within the response length and keeps the distances between later queries and
the codes close to those of the original block (properties (i) and (iii) below). The per-layer budgets of
Section~\ref{sec:allocation} add one requirement: a layer that keeps only its first $q$ pause tokens should still see codes
spread over the whole block, not over its first $q\,C/K_{\max}$ tokens. Bit-reversing the order in which the evenly spaced
positions are assigned achieves this:
\begin{equation}\label{eq:pos}
\mathrm{pos}(x_{b,t}) = P + bC + t, \qquad
\mathrm{pos}(p_{b,k}) = P + bC + \mathrm{bitrev}_{K_{\max}}(k)\,\frac{C}{K_{\max}},
\end{equation}
where $\mathrm{bitrev}_{K_{\max}}(k)$ reverses the $\log_2 K_{\max}$ bits of $k$ (for $K_{\max}{=}8$ the slot order is
$0,4,2,6,1,5,3,7$), so the first $q$ pause tokens always sit on an even grid of stride $C/q$. Query and key use the same
rotation, so this is a choice of position ids only. Three properties follow.
\emph{(i) No position inflation.} A response of $T$ tokens occupies exactly $T$ positions whatever $K_{\max}$ is, so a 32K-token
response stays inside the $32{,}768$ positions the base model was trained on. Giving the codes fresh positions after their block,
as the learned-summary literature does, stretches the same response over $T(C{+}K)/C$ positions.
\emph{(ii) Nested, $K$-independent quotas.} Keeping the first $q$ slots in bit-reversed order, $k<q$, keeps an even grid of stride
$C/q$ over the block (Figure~\ref{fig:layout}b): every quota level sees an evenly spaced summary, quota $q$ is a prefix of quota
$2q$, and the layout of the kept rows does not depend on $K_{\max}$, so a per-(layer, head) budget is a mask on one trained model
rather than a different model. \emph{(iii) Stable relative geometry.} A later query sees each code at roughly the distance the
content it replaces would have had, rather than at a distance that grows with the number of intervening summaries. Ablating only the in-block placement (giving the code rows new positions after their block), with the mask, budget, trainer
and data fixed, lowers accuracy by $6.2$ points of avg@8 (95\% interval $[4.6, 7.8]$) at the 32K cap; the loss is not caused by
positions beyond the base model's trained range (Section~\ref{sec:ablations}). The in-block layout has one cost. A pause token's query sits at its in-block position, so the tokens of its block that come
after that position are, for this query, at negative relative offsets: keys positioned after the query. The pretrained weights
have never seen such offsets, so fine-tuning has to learn to use them.

\paragraph{Quotas.} Each layer $\ell$ keeps only the first $q_\ell\le K_{\max}$ code rows of every completed block and masks
the rest; at serving time only those rows persist (Appendix~\ref{app:serving}). Section~\ref{sec:allocation} sets the quotas.

\subsection{Training objective}
The model is trained with the ordinary next-token loss on every response token, computed while its history is held as
codes (Eq.~\ref{eq:ce}); the prompt is never compressed. The whole response, code rows included, is one forward pass under the
block mask, so the loss of later blocks flows back into the pause tokens that wrote the codes they read
(Figure~\ref{fig:layout}d). We add no auxiliary terms: distillation from a frozen full-context teacher and keeping a raw tail
were tried and did not help (Appendix~\ref{app:archived}).
\begin{equation}\label{eq:ce}
\mathcal{L}_{\mathrm{CE}} = -\tfrac{1}{|R|}\sum_{t\in R}\log p_\theta\!\left(x_t \mid \text{prompt},\ \text{codes}_{<b(t)},\ x_{b(t),<t}\right)
\end{equation}

\subsection{Allocating the budget across layers}
\label{sec:allocation}
The budget is a per-layer quota $q_\ell \le K_{\max}$: layer $\ell$ keeps $q_\ell$ of the $K_{\max}{=}256$ code rows of every
completed block and masks the rest, the same quota for every KV head of the layer (so tensor-parallel serving stays balanced and
the physical cache can drop the masked rows, Appendix~\ref{app:serving}). \emph{Uniform} allocation sets $q_\ell{=}128$ everywhere.
\emph{Per-layer} (dynamic) allocation keeps the same mean of 128 rows but moves rows between layers, in three steps:
\begin{enumerate}\itemsep0pt
  \item Train a model with the uniform-128 schedule.
  \item Measure what each layer's code rows are worth: at evaluation time, without retraining, halve one layer's quota
  (128 $\to$ 64) and record the increase in held-out cross-entropy; repeat for each of the 36 layers (Figure~\ref{fig:budget}a).
  Most layers cost nothing; a few cost a lot.
  \item Rank layers by cost: the 6 most expensive get 256 rows, the 12 cheapest 64, the other 18 keep 128
  ($6{\cdot}256 + 18{\cdot}128 + 12{\cdot}64 = 36{\cdot}128$), and train a new model on this schedule.
\end{enumerate}
The resulting schedule (Figure~\ref{fig:budget}) raises layers 0, 15, 18, 20, 21 and 24 and cuts layers 1--6, 8, 10,
30--32 and 35. Section~\ref{sec:analysis} explains what the measurement shows, why we use three levels and cut only a third of
the layers, and why the quota is set per layer rather than per head.

\section{Experimental setup}
\label{sec:setup}
Every arm fine-tunes Qwen3-4B-Base on the same 2B response tokens of math SFT data (400K prompt/response pairs derived from
OpenMathReasoning) in the same canonical order, with $C{=}512$, $K_{\max}{=}256$ slots and a mean budget of 128 code rows per
block unless stated; each arm is served with the schedule it was trained with. Arms are named by their block-to-code ratio (this
budget is 4:1; 2:1 and 8:1 keep 256 and 64 rows) and their allocation, uniform or per-layer; \emph{appended positions} is the ablation whose code rows take new
positions after the block. Full hyperparameters, hardware and throughput are in
Appendix~\ref{app:train}. The reference is the same base model fine-tuned on the same tokens with full KV, the ceiling a compressed model can
match but not beat; every arm is reported as a gap to it.

\paragraph{Evaluation.} A 446-problem competition suite (twelve competitions from AIME, HMMT, BRUMO, CMIMC, SMT and AMC, 2022--2026;
Appendix~\ref{app:suite}), 8 samples per problem at $T{=}0.6$, top-$p$ 0.95, top-$k$ 20, seed 1234, and a 32K cap (31{,}744
generated tokens); a sample is correct if its last boxed answer matches or math-verify judges the full text equivalent. Every arm
is scored on the same problems and seed, so differences are paired per problem, with a 2000-resample per-problem bootstrap.
Pooling makes these comparisons resolvable: on one 30-problem set the 95\% interval on a paired gap is about $\pm5$ points,
and a full-KV re-run differing only in its data seed (the same data pool, with a different random draw of which examples
each training step uses, and in what order) scores significantly below the original full-KV model on AIME 2024 and above it
on AIME 2025; on the pooled suite the half-width is $1.5$
(Appendix~\ref{app:pooled}). We report avg@8, the mean accuracy over the 8 samples of each problem; MATH-500 is scored as avg@4 at a 16K cap. We measure
budget as the number of persistent KV rows kept per 512 generated tokens per KV head (``KV/512'' in Table~\ref{tab:main}).

\paragraph{Serving.} Speed is measured in vLLM on one H100 with our patch for compressed decoding. A completed block's code rows
are computed in the next engine step alongside other requests' decode tokens and its raw rows are released; a per-layer schedule
is stored as one KV cache group per quota level, so rows a layer does not keep occupy no memory. Every arm runs under CUDA graphs
and vLLM's asynchronous scheduling, except R-KV, whose port applies each compaction before the next step, and LightThinker
(Appendix~\ref{app:lightthinker}). Time to
finish is measured by replaying 512 requests at each arm's own suite lengths, run to completion at the concurrency cap that
finishes fastest. The served model is the
trained one: vLLM greedy tokens match the reference decoder, and trainer and incremental generator agree on per-token CE to
$4\times10^{-4}$ (Appendix~\ref{app:parity}). Implementation and per-arm measurements are in Appendix~\ref{app:serving}.

\section{Results}
\label{sec:results}
\begin{figure}[t]
  \centering
  \includegraphics[width=\linewidth]{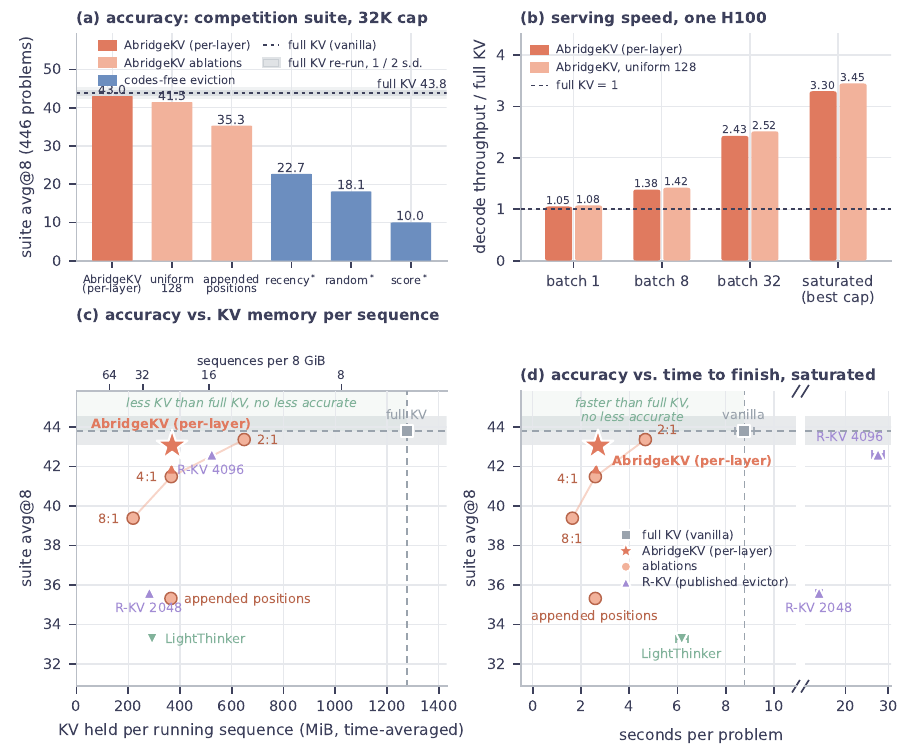}
  \caption{Main results. Orange circles are our arms and the orange star the per-layer model; grey squares full KV, purple
  triangles R-KV, green LightThinker; grey band: 1 s.d.\ of a full-KV re-run. (a)~Suite avg@8, 32K cap ($^*$60-problem subset).
  (b)~Decode throughput relative to full KV on one H100, for sequences of 16K tokens. (c)~Accuracy against KV held per running sequence, time-averaged
  over each arm's generations; the arrow marks the accuracy gain of the per-layer schedule over the uniform one at the same memory. (d)~Accuracy against time to
  finish a problem: 512 suite-length requests run to completion at the fastest concurrency cap; shaded: at least as accurate
  as full KV and faster; the axis is broken to show R-KV; its horizontal bar extends to the time R-KV would take if computing its eviction scores took no time.
  Other horizontal bars: 95\% intervals on mean generated length.}
  \label{fig:main}\label{fig:pareto}
\end{figure}
\begin{table}[t]
\centering\footnotesize\setlength{\tabcolsep}{2.6pt}
\caption{Main comparison at the 32K cap. Suite: avg@8 over 446 problems with 95\% bootstrap intervals; $\Delta$: paired
difference to the full-KV model (vanilla). KV/512: persistent KV rows per 512 generated tokens. trunc.: share of suite samples
reaching the cap. $^\dagger$~Eviction baselines, run on a 60-problem subset (AIME 2025 and HMMT Feb 2025) on which full KV scores 29.8 and
the per-layer arm 28.3: their $\Delta$ is paired on those problems and comparable with the other rows' $\Delta$, their avg@8 is not.
Protocol details: Appendix~\ref{app:suite}.
Shaded: ours; bold: best below the reference rows.}
\label{tab:main}
\begin{tabular}{@{}lrrrllr@{}}
\toprule
Arm & KV/512 & CE & MATH-500 & Suite avg@8 & $\Delta$ vs.\ vanilla & trunc.\% \\
\midrule
\multicolumn{7}{l}{\textit{References}}\\
Base model (no SFT) & 512 & -- & 69.6 & 13.5 {\scriptsize[11.0, 16.2]} & -30.4 {\scriptsize[-33.4, -27.0]} & 5.6 \\
Vanilla SFT (full KV) & 512 & 0.4162 & 88.1 & 43.8 {\scriptsize[39.8, 47.6]} & -- & 3.4 \\
\midrule
\multicolumn{7}{l}{\textit{Ours: dynamic per-layer allocation}}\\
\rowcolor{oursbg}\method{} (per-layer) & 128 & \textbf{0.4215} & 87.5 & \textbf{43.0} {\scriptsize[39.1, 46.9]} & \textbf{-0.8} {\scriptsize[-2.3, +0.8]} & \textbf{3.2} \\
\midrule
\multicolumn{7}{l}{\textit{Ablations: one design choice removed}}\\
\rowcolor{ablbg}uniform allocation & 128 & 0.4227 & \textbf{88.3} & 41.5 {\scriptsize[37.6, 45.3]} & -2.3 {\scriptsize[-3.8, -0.9]} & 3.4 \\
\rowcolor{ablbg}appended positions & 128 & 0.4297 & 85.8 & 35.3 {\scriptsize[31.6, 39.0]} & -8.5 {\scriptsize[-10.3, -6.8]} & 3.6 \\
\midrule
\multicolumn{7}{l}{\textit{Prior learned compression, in our training and serving stack}}\\
LightThinker (thought-level) & 97 & 0.4391 & 83.4 & 33.3 {\scriptsize[29.6, 37.0]} & -10.5 {\scriptsize[-12.5, -8.7]} & 5.4 \\
\midrule
\multicolumn{7}{l}{\textit{Training-free eviction (vanilla weights)}}\\
sink + recency & 128 & -- & 84.0 & 22.7 {\scriptsize[14.6, 32.3]}$^\dagger$ & -7.1 {\scriptsize[-10.4, -3.8]}$^\dagger$ & 37.9 \\
random & 128 & -- & 76.1 & 18.1 {\scriptsize[11.0, 26.2]}$^\dagger$ & -11.7 {\scriptsize[-16.5, -7.3]}$^\dagger$ & 52.9 \\
attention score (H2O) & 128 & -- & 59.1 & 10.0 {\scriptsize[5.0, 15.8]}$^\dagger$ & -19.8 {\scriptsize[-26.2, -13.3]}$^\dagger$ & 57.7 \\
R-KV, budget 2048 & 74 & -- & 83.9 & 35.6 {\scriptsize[31.7, 39.5]} & -8.2 {\scriptsize[-10.2, -6.3]} & 55.8 \\
R-KV, budget 4096 & 144 & -- & 86.5 & 42.6 {\scriptsize[38.5, 46.6]} & -1.2 {\scriptsize[-2.7, +0.3]} & 31.7 \\
\bottomrule
\end{tabular}
\end{table}

\subsection{Accuracy and generation length}
\label{sec:parity}
At a quarter of the persistent cache, \method{} reaches $43.0$ avg@8 against $43.8$ for full KV (Table~\ref{tab:main},
Figure~\ref{fig:main}a), a paired gap of $-0.8$
$[-2.3, +0.8]$ that lies inside the spread of a full-KV re-run differing only in its data seed (1 s.d.\ 0.8 points, Appendix~\ref{app:archived}). On MATH-500, whose solutions are short, the two are $87.5$ and $88.1$.

\begin{figure}[t]
  \centering
  \includegraphics[width=\linewidth]{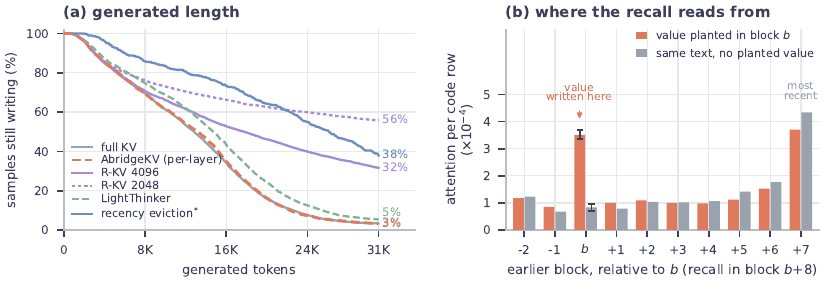}
  \caption{(a)~Share of samples still writing at each length, and truncated at the cap ($^*$60-problem subset).
  (b)~A value planted in block $b$ (``$Q_7 = 7391$'') is recalled 8 compressions later. The plot shows the attention that the recalled digits pay to each
  earlier block's code rows, per code row, averaged over heads and layers, with and without the planted value; 93 edited
  solutions, with a 95\% interval at block $b$.}
  \label{fig:lenflow}
\end{figure}

\paragraph{Generation length.} Matching accuracy could hide a model that writes more to make up for
what it has lost. \method{} does not: its length distribution is statistically indistinguishable from full KV's, as close as
a full-KV re-run that differs only in its data seed. The Kolmogorov--Smirnov distance from full KV's length distribution is
0.024 for \method{} and 0.017 for the re-run; at our sample size a distance of 0.032 would be significant at the 5\% level, so
neither is. The same problems also run long under both: the per-problem correlation of generated length with full KV is 0.93
for \method{} and 0.92 for the re-run (Figure~\ref{fig:lenflow}a). R-KV at budget 4096, the budget matched to our 4:1 arm, is far from both: the Kolmogorov--Smirnov distance between its length
distribution and full KV's is 0.33, and 32\% of its samples run to the 32K cap.

\subsection{Serving efficiency}
\label{sec:serving}
\method{}'s resident KV per sequence is $3.54\times$ smaller than full KV's (the ceiling is $4\times$; the prompt and the open
block stay raw), and because the per-layer schedule is a handful of cache groups on the stock paged path, it decodes as fast as
the uniform setting: $1.05\times$, $1.38\times$ and $2.43\times$ full KV's throughput at batch 1, 8 and 32
(Figure~\ref{fig:main}b).
Averaged over a whole generation, a running \method{} sequence holds 369~MiB of KV, against 1{,}277~MiB for full KV. This is
the same memory as the uniform setting, while \method{}'s accuracy is 1.6 points higher. At these sizes an 8~GiB cache holds 22
\method{} sequences instead of 6 full-KV sequences
(Figure~\ref{fig:main}c). Figure~\ref{fig:main}d then charges each arm for what it generates: saturated, \method{} finishes a
problem in $0.31\times$ full KV's time, the same as the uniform setting ($0.30\times$). It is the only compressed arm inside the shaded region of Figure~\ref{fig:main}d, which contains arms at least as accurate as
full KV and faster than it. R-KV (Section~\ref{sec:evict}) takes longer than full KV to finish a problem, because it generates
more tokens and runs a compaction every 64 decode steps.

\subsection{Comparison with eviction}
\label{sec:evict}
Training-free eviction at the same budget does not come close, which is expected for two reasons. Eviction selects entries
by the attention they have received so far, but a CoT reuses intermediate results at later points that past attention does not
predict; and an evicted entry leaves no representation, whereas a learned code row retains a summary of every block in a form
the model is trained to read. When a needed result has been evicted, the model tends to recompute it, and generations lengthen.

Applied to the full-KV model, eviction that keeps 128 of each completed block's 512 rows, selected by recency, at random, or by
accumulated attention score, loses 7--20 points of avg@8 against full KV. These three policies were run only on a 60-problem
subset (AIME 2025 and HMMT Feb 2025), and the differences are paired on those problems. The truncation column of
Table~\ref{tab:main} shows the mechanism: 38--58\% of their samples run to the 32K cap. R-KV~\citep{cai2025rkv}, the published evictor
designed for long reasoning, which we port to our serving stack and run on the same checkpoint, problems, sampling and grader, is
the strongest of the family. We run it in its servable form, one kept set shared by all heads as in its authors' own SGLang
port, since a paged cache cannot keep different tokens per head (Appendix~\ref{app:rkv}). We compare each R-KV budget with the arm whose cache it most nearly matches, giving R-KV the slightly larger cache: a fixed
budget of 4096 rows leaves about 12\% more rows resident than our 4:1 arm, and 2048 about 12\% more than our 8:1 arm
(Appendix~\ref{app:rkv}). At budget 4096, R-KV's accuracy is within noise of \method{} at 4:1 (paired difference $-0.4$
$[-2.1, +1.2]$ avg@8 points). At budget 2048, it is below our 8:1 arm ($-3.8$ $[-5.5, -2.2]$). Like the simple eviction policies, R-KV lengthens the output: $32\%$ of its samples reach the cap at budget 4096 and $56\%$ at
budget 2048, and it generates 40--69\% more tokens than full KV on average (40\% at budget 4096, 69\% at 2048;
Figure~\ref{fig:lenflow}a), so the memory it saves per token is spent on extra tokens and it takes longer than full KV to finish a problem
(Figure~\ref{fig:main}d). The full comparison, including the port's parity check, is in Appendix~\ref{app:rkv}.

\subsection{Ablations}
\label{sec:ablations}
The ablation rows of Table~\ref{tab:main} remove one design choice at a time, holding trainer, data, order, block size and mean
budget fixed; Appendix~\ref{app:scaling} gives the full block-size and budget sweep.

\paragraph{Code positions.} Giving each code row a fresh position after its block, the convention
of learned summaries~\citep{zhang2024activationbeacon,chevalier2023autocompressors,mu2023gist,zhang2025lightthinker}, lowers avg@8 by $6.2$ $[4.6, 7.8]$ points relative to the uniform arm with in-block positions, the largest effect we measure
(the $-8.5$ in Table~\ref{tab:main} is its gap to full KV). The loss is not caused by position inflation. With appended
positions, a response exceeds the positions the base model was trained on only after $26{,}214$ generated tokens, and $6.3\%$ of
samples reach that length. Restricting the comparison to the 256 problems on which no sample of either arm reaches it leaves the
gap unchanged at $6.2$ $[4.1, 8.3]$. An in-block row keeps the distance to later queries that its content would have had
(Appendix~\ref{app:positions}).

\paragraph{Per-layer allocation.} Spending the mean of 128 rows uniformly across layers loses $2.3$ $[0.9, 3.8]$ points against full KV. Allocating the same
mean by each layer's measured value (Section~\ref{sec:allocation}) recovers $1.6$ $[0.1, 3.0]$ of these points, measured as a
paired difference between the per-layer and uniform arms, and leaves a gap to full KV ($-0.8$ $[-2.3, +0.8]$) whose interval
contains zero.\ifarxiv\ The difference is positive on every subset of the suite we examined, but no subset on its own is large enough
to resolve a difference of this size (Appendix~\ref{app:suite}).\fi

\begin{figure}[t]
  \centering
  \includegraphics[width=\linewidth]{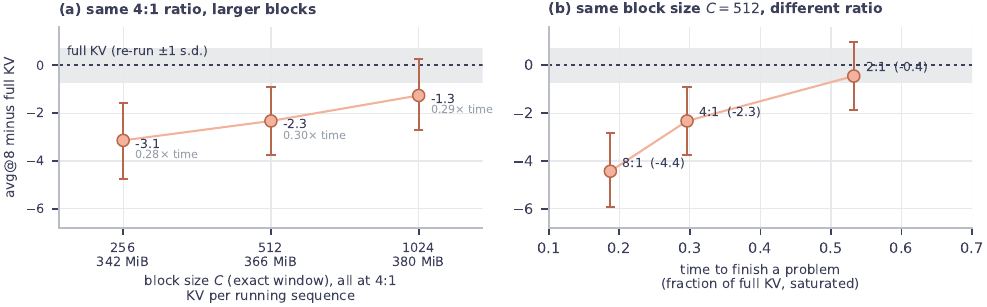}
  \caption{Compression shape; paired gap to full KV with 95\% intervals (grey: 1 s.d.\ of a full-KV re-run). (a)~At the same 4:1 ratio, larger blocks hold slightly more KV per sequence (second row of tick labels) and go through fewer
  lossy compressions; accuracy rises and time to finish (grey) stays the same. (b)~At the same block size, a higher ratio is
  faster and less accurate.}
  \label{fig:cksweep}\label{fig:blockscale}
\end{figure}
\paragraph{Block size at a fixed ratio.} A larger block keeps a longer window of recent tokens exact, which costs memory, and lowers the number of lossy compressions
the history passes through: at 4:1, a query at our mean generation length comes after 51 compressed blocks at $C{=}256$ and 13
at $C{=}1024$. Suite accuracy increases with block size: $40.7$, $41.5$ and $42.5$ avg@8 for $C{=}256$, $512$ and $1024$
(Figure~\ref{fig:cksweep}a). Over this range the KV held per running sequence grows by $11\%$ (342 to 380~MiB), while time to
finish does not change measurably ($0.28$, $0.30$ and $0.29\times$ full KV's). Only the smallest block is significantly below
full KV: the gap is $-3.1$ $[-4.7, -1.6]$ at $C{=}256$ and $-1.3$ $[-2.7, +0.3]$ at $C{=}1024$. Thought-level compression sits far to the small side of this range. LightThinker's algorithm, trained on our data and served
in our engine, ends a thought at each paragraph break, every 53 tokens on average in its generations, and compresses the thought into 10
rows (9 cache tokens and its continue token), a ratio of about 5.3:1. It scores
$33.3$ on the suite, $-10.5$ $[-12.5, -8.7]$ against full KV (Table~\ref{tab:main}; Appendix~\ref{app:lightthinker}); our
8:1 arm, which keeps fewer rows, loses $4.4$. It is also slower: it takes $0.70\times$ full KV's time to finish a problem, against $0.31\times$ for \method{}
(Figure~\ref{fig:main}d), since its compression step runs ten times as often. Its measured time is an upper bound, because it
ran with synchronous scheduling (Appendix~\ref{app:lightthinker}). Training cost does not go up with block size over this range (Appendix~\ref{app:blockcost}).

\paragraph{Compression ratio at a fixed block size.} At $C{=}512$, the 2:1 arm is statistically indistinguishable from full KV ($-0.4$ $[-1.8, +1.0]$) and finishes a problem in
$0.53\times$ full KV's time. With uniform allocation, the 4:1 arm loses $2.3$ points and takes $0.30\times$ full KV's time, and
the 8:1 arm loses $4.4$ points and takes $0.19\times$ (Figure~\ref{fig:cksweep}b). The per-layer schedule brings the 4:1 arm
to parity with full KV at the same time to finish.

\section{Analysis}
\label{sec:dynamic}
\label{sec:analysis}
\subsection{Choosing the quotas}
\label{sec:quotas}
\begin{figure}[t]
  \centering
  \includegraphics[width=\linewidth]{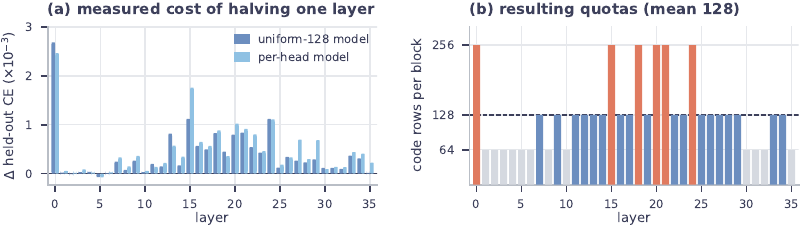}
  \caption{From measurement to quotas. (a)~Held-out CE cost of halving one layer's code rows at evaluation time, on two trained
  models. (b)~The trained schedule: the six most expensive layers get 256 rows, the twelve cheapest 64, at the same mean.}
  \label{fig:budget}
\end{figure}
\paragraph{Question.} Allocating rows by layer turns a significant loss into parity (Section~\ref{sec:ablations}). We want to
know whether the gain comes from \emph{where} the rows go, which would make the measurement behind the schedule essential, or
merely from making the budget uneven.

\paragraph{Procedure.} On the trained uniform model we halve one layer's code rows at evaluation time and record the increase
in held-out cross-entropy, for each of the 36 layers (Figure~\ref{fig:budget}a). The most expensive layers are raised, the
cheapest cut, at the same mean, and the model is retrained on that schedule (Figure~\ref{fig:budget}b). For contrast we train
a schedule of the same mean chosen without the trained model, from the heads whose full-KV attention is hardest to
approximate with few keys; it raises the early layers.

\paragraph{Result.} The measured cost sits in a few layers, the first layer and a band in the middle of the network, while
most early and late layers can lose half their rows at almost no cost. Moving rows by this measurement reaches parity with full
KV; the schedule chosen from the full-KV model's attention, which raises exactly the early layers the measurement finds cheap,
is no better than uniform (Appendix~\ref{app:dynamic}, Table~\ref{tab:allocation}). The placement is what matters, and it has to be read from the trained compressed model: a proxy
from the uncompressed model points the wrong way. Two refinements did not improve on this schedule. First, the single-layer costs are not additive: halving every layer at once
raises held-out cross-entropy by twice the sum of the single-layer increases, so rather than fitting small differences between
layers, we cut only a third of the layers and use three quota levels. Second, a per-head schedule matches the per-layer accuracy
but costs memory or speed to serve (Appendix~\ref{app:dynamic}).

\subsection{Information flow through the code rows}
\label{sec:probe}
\paragraph{Question.} Once a block is compressed its raw KV is gone, so anything later reasoning takes from that block must
pass through its code rows. We want to see this directly: whether a specific fact written in a block is stored in that
block's code rows and retrieved from them many blocks later.

\paragraph{Procedure.} We insert one sentence defining a random four-digit value (``Denote this intermediate result as $Q_7 =
7391$.'') into block $b$ of the per-layer model's own AIME solutions, ask for it $k$ blocks later (``Recall that $Q_7 =$''), and
record whether it is reproduced and where its tokens attend.

\paragraph{Result.} Recall is scored by teacher forcing: after the cue, the value counts as recalled if the model's most likely
next tokens are exactly its four digits, so each case is a deterministic yes or no rather than a sample. Over 93 edited
solutions from 51 AIME problems, each asked at 1, 2, 4 and 8 compressions after block $b$ (372 cases), the value is recalled in
$73\%$ of cases ($95\%$ interval $66$--$79\%$), against $98\%$ for the full-KV model on the same edited sequences. The rate
does not decline from one to eight compressions, and it is zero when block $b$'s code rows are removed. The recalling tokens attend to
block $b$'s code rows far more than to any other block's and far more than the same positions do when no value was planted
(Figure~\ref{fig:lenflow}b): the information travels by the most direct route, from the block into its own code rows and from
there to the token that needs it. This routing happens in the 256-row layers, the ones the measurement of
Section~\ref{sec:quotas} raised, and in the 128-row layers. Failed recalls attend to the right rows as strongly as successful ones, so what limits recall
is how much the code rows hold, not whether the model finds them. The same holds without a planted value: in the model's own solutions, when the next block reuses a number first written in block
$b$, the probability the model assigns to the number's later digits is $0.99$ with block $b$'s code rows intact and about $0.69$
when they are replaced by another block's rows or removed (Figure~\ref{fig:probe}, Appendix~\ref{app:probe}).

\section{Discussion}
\label{sec:discussion}
\paragraph{Scope.} Mathematical reasoning is the common test bed of the closest related
work~\citep{kontonis2026memento,zhang2025lightthinker,cai2025rkv}, and it is ours. It is a demanding test of long generation:
solutions run to tens of thousands of tokens, correctness is verifiable, and our suite pools twelve competitions over five years,
446 problems, so the claims above are resolved at a precision single benchmarks do not reach. Nothing in the recipe is specific
to the domain or the model: the trainer, the layout and the allocation procedure take any decoder with rotary positions, the
allocation is re-measured on each trained model rather than transferred, and the serving path is the stock paged engine with
one cache group per quota. \ifarxiv A first coding-domain comparison at prototyping scale (Appendix~\ref{app:code}) shows the same ordering of full KV,
per-layer and uniform, and a per-layer schedule measured on code that coincides with the one measured on mathematics; larger
models may also be more compressible, since their KV states carry more redundancy per token.\fi

\paragraph{Block size.} Accuracy is still rising at $C{=}1024$, so larger blocks may be better still; we did not train them,
although they cost little more to train (Appendix~\ref{app:blockcost}). The trade-off is memory: a larger block keeps more raw
rows resident per sequence, so at a fixed cache it means fewer sequences in flight. We expect the practical choice to be the
largest block the deployment's cache can carry.

\paragraph{Readable reasoning and other modalities.} Reasoning in a latent space instead of text~\citep{hao2024coconut}
shortens generations but gives up the readable chain of thought that makes a model's reasoning possible to
monitor~\citep{korbak2025monitorability}. \method{} keeps the reasoning in text: every token the model writes is ordinary text
that can be read and checked, and only the cache that later tokens attend to is compressed. Because the summaries are learned
from the next-token loss rather than written out as text, the same recipe also applies where a faithful text summary is hard to
define, such as long contexts of images or other non-text inputs, whose KV could be compressed in the same way.

\paragraph{Limitations.} Every result starts from a base model fine-tuned with our own SFT. This matches how reasoning models
are commonly built, supervised fine-tuning on long reasoning traces followed by reinforcement
learning~\citep{abdin2025phi4reasoning}, and compression training fits into that SFT stage. It fits equally into the
distillation step of pipelines that train separate domain-specific specialists with RL and then distill them into one
model~\citep{deepseek2025v32}. We did not train it directly on a released RL-tuned checkpoint, and whether further fine-tuning
there preserves the gains of RL is untested. The compression ratio is fixed per block, so a block that carries little and one that carries a lot get the same
rows. And the quota schedule comes from single-layer measurements that we assumed to be independent
(Section~\ref{sec:quotas}); a method that measured how cuts interact could find a better one.

\section*{Reproducibility statement}
Every arm is trained from Qwen3-4B-Base on the same 2B response tokens in the same order (the canonical data stream), with
the hyperparameters, hardware and throughput of Appendix~\ref{app:train}; the 446-problem suite, its sources and the sampling
protocol are in Section~\ref{sec:setup} and Appendix~\ref{app:suite}; the serving implementation, the benchmark protocol
(pinned CPU cores, CUDA graphs on every arm, the shared concurrency-cap grid) and the per-arm measurements are in
Appendix~\ref{app:serving}; and the parity checks that tie the served model to the trained one are in
Appendix~\ref{app:parity}. We will release the training code, the vLLM patch, the evaluation and benchmark scripts, and the
trained checkpoints.

\section*{AI use statement}
In this work, we used generative AI tools (a coding agent) for implementing methods, including the training, evaluation,
serving and benchmark code; for running and monitoring experiments; for drafting and editing text; for creating figures and
tables from the logged results; and for searching and summarizing the literature. We have not used generative AI tools for
research ideas or for interpreting results, and we have not used them for generating synthetic datasets: our SFT data are
derived from \ifarxiv OpenMathReasoning and OpenCodeReasoning, public datasets\fi{} of model-written solutions that we did not generate. Developing theoretical
models, formulating mathematical claims, providing or writing proofs, translation, and qualitative data analysis are not
applicable to this work. Every reported number was produced by code we ran and was checked against the underlying logs. We
take responsibility for the final content of this work, including text, claims or artifacts produced with the aid of
generative AI.

\bibliography{main}

\begin{thebibliography}{62}
\providecommand{\natexlab}[1]{#1}
\providecommand{\url}[1]{\texttt{#1}}
\expandafter\ifx\csname urlstyle\endcsname\relax
  \providecommand{\doi}[1]{doi: #1}\else
  \providecommand{\doi}{doi: \begingroup \urlstyle{rm}\Url}\fi

\bibitem[Abdin et~al.(2025)Abdin, Agarwal, Awadallah, Balachandran, Behl, Chen,
  et~al.]{abdin2025phi4reasoning}
Marah Abdin, Sahaj Agarwal, Ahmed Awadallah, Vidhisha Balachandran, Harkirat
  Behl, Lingjiao Chen, et~al.
\newblock {Phi-4-reasoning} technical report.
\newblock \emph{arXiv preprint arXiv:2504.21318}, 2025.

\bibitem[Adnan et~al.(2024)Adnan, Arunkumar, Jain, Nair, Soloveychik, and
  Kamath]{adnan2024keyformer}
Muhammad Adnan, Akhil Arunkumar, Gaurav Jain, Prashant~J. Nair, Ilya
  Soloveychik, and Purushotham Kamath.
\newblock Keyformer: Kv cache reduction through key tokens selection for
  efficient generative inference.
\newblock In \emph{Proceedings of Machine Learning and Systems (MLSys)}, 2024.

\bibitem[Aghajohari et~al.(2025)Aghajohari, Chitsaz, Kazemnejad, Chandar,
  Sordoni, Courville, and Reddy]{aghajohari2025markovian}
Milad Aghajohari, Kamran Chitsaz, Amirhossein Kazemnejad, Sarath Chandar,
  Alessandro Sordoni, Aaron Courville, and Siva Reddy.
\newblock The {Markovian} thinker: Architecture-agnostic linear scaling of
  reasoning.
\newblock \emph{arXiv preprint arXiv:2510.06557}, 2025.

\bibitem[Ahmad et~al.(2025)Ahmad, Narenthiran, Majumdar, Ficek, Jain, Huang,
  Noroozi, and Ginsburg]{ahmad2025opencodereasoning}
Wasi~Uddin Ahmad, Sean Narenthiran, Somshubra Majumdar, Aleksander Ficek,
  Siddhartha Jain, Jocelyn Huang, Vahid Noroozi, and Boris Ginsburg.
\newblock {OpenCodeReasoning}: Advancing data distillation for competitive
  coding.
\newblock \emph{arXiv preprint arXiv:2504.01943}, 2025.

\bibitem[Ainslie et~al.(2023)Ainslie, Lee-Thorp, de~Jong, Zemlyanskiy,
  Lebr{\'o}n, and Sanghai]{ainslie2023gqa}
Joshua Ainslie, James Lee-Thorp, Michiel de~Jong, Yury Zemlyanskiy, Federico
  Lebr{\'o}n, and Sumit Sanghai.
\newblock Gqa: Training generalized multi-query transformer models from
  multi-head checkpoints.
\newblock \emph{arXiv preprint arXiv:2305.13245}, 2023.
\newblock EMNLP 2023.

\bibitem[Brandon et~al.(2024)Brandon, Mishra, Nrusimha, Panda, and
  Ragan-Kelley]{brandon2024cla}
William Brandon, Mayank Mishra, Aniruddha Nrusimha, Rameswar Panda, and
  Jonathan Ragan-Kelley.
\newblock Reducing transformer key-value cache size with cross-layer attention.
\newblock \emph{arXiv preprint arXiv:2405.12981}, 2024.
\newblock NeurIPS 2024.

\bibitem[Cai et~al.(2024)Cai, Zhang, Gao, Liu, Liu, Lu, Xiong, Dong, Chang, Hu,
  and Xiao]{cai2024pyramidkv}
Zefan Cai, Yichi Zhang, Bofei Gao, Yuliang Liu, Tianyu Liu, Keming Lu, Wayne
  Xiong, Yue Dong, Baobao Chang, Junjie Hu, and Wen Xiao.
\newblock Pyramidkv: Dynamic kv cache compression based on pyramidal
  information funneling.
\newblock \emph{arXiv preprint arXiv:2406.02069}, 2024.

\bibitem[Cai et~al.(2025)Cai, Xiao, Sun, Luo, Zhang, Wan, Li, Zhou, Chang, Gu,
  Dong, Anandkumar, Asi, and Hu]{cai2025rkv}
Zefan Cai, Wen Xiao, Hanshi Sun, Cheng Luo, Yikai Zhang, Ke~Wan, Yucheng Li,
  Yeyang Zhou, Li-Wen Chang, Jiuxiang Gu, Zhen Dong, Anima Anandkumar,
  Abedelkadir Asi, and Junjie Hu.
\newblock R-kv: Redundancy-aware kv cache compression for reasoning models.
\newblock In \emph{Advances in Neural Information Processing Systems}, 2025.

\bibitem[Chang et~al.(2026)Chang, Fu, Fu, Yang, Thomason, and
  Jia]{chang2026vase}
Ting-Yun Chang, Harvey~Yiyun Fu, Deqing Fu, Chenghao Yang, Jesse Thomason, and
  Robin Jia.
\newblock Value-aware stochastic kv cache eviction for reasoning models.
\newblock \emph{arXiv preprint arXiv:2606.03928}, 2026.

\bibitem[Cheng \& Van~Durme(2024)Cheng and Van~Durme]{cheng2024ccot}
Jeffrey Cheng and Benjamin Van~Durme.
\newblock Compressed chain of thought: Efficient reasoning through dense
  representations.
\newblock \emph{arXiv preprint arXiv:2412.13171}, 2024.

\bibitem[Chevalier et~al.(2023)Chevalier, Wettig, Ajith, and
  Chen]{chevalier2023autocompressors}
Alexis Chevalier, Alexander Wettig, Anirudh Ajith, and Danqi Chen.
\newblock Adapting language models to compress contexts.
\newblock In \emph{Proceedings of EMNLP}, pp.\  3829--3846, 2023.

\bibitem[{DeepSeek-AI}(2024)]{deepseek2024v2}
{DeepSeek-AI}.
\newblock Deepseek-v2: A strong, economical, and efficient mixture-of-experts
  language model.
\newblock \emph{arXiv preprint arXiv:2405.04434}, 2024.

\bibitem[{DeepSeek-AI}(2025{\natexlab{a}})]{deepseek2025r1}
{DeepSeek-AI}.
\newblock Deepseek-r1 incentivizes reasoning in llms through reinforcement
  learning.
\newblock \emph{Nature}, 645\penalty0 (8081):\penalty0 633--638,
  2025{\natexlab{a}}.
\newblock \doi{10.1038/s41586-025-09422-z}.

\bibitem[{DeepSeek-AI}(2025{\natexlab{b}})]{deepseek2025v32}
{DeepSeek-AI}.
\newblock {DeepSeek-V3.2}: Pushing the frontier of open large language models.
\newblock \emph{arXiv preprint arXiv:2512.02556}, 2025{\natexlab{b}}.

\bibitem[Feng et~al.(2025)Feng, Lv, Cao, Xie, and Zhou]{feng2025adakv}
Yuan Feng, Junlin Lv, Yukun Cao, Xike Xie, and S.~Kevin Zhou.
\newblock Ada-kv: Optimizing kv cache eviction by adaptive budget allocation
  for efficient llm inference.
\newblock In \emph{Advances in Neural Information Processing Systems}, 2025.

\bibitem[Fu et~al.(2025)Fu, Cai, Asi, Xiong, Dong, and Xiao]{fu2025headkv}
Yu~Fu, Zefan Cai, Abedelkadir Asi, Wayne Xiong, Yue Dong, and Wen Xiao.
\newblock Not all heads matter: A head-level kv cache compression method with
  integrated retrieval and reasoning.
\newblock In \emph{International Conference on Learning Representations
  (ICLR)}, 2025.

\bibitem[Gao et~al.(2025)Gao, Guo, Cao, Xia, Cheng, Wang, Ma, Sun, Ye, Dong,
  So, Hua, Cao, Yang, and Yang]{gao2025seerattentionr}
Yizhao Gao, Shuming Guo, Shijie Cao, Yuqing Xia, Yu~Cheng, Lei Wang, Lingxiao
  Ma, Yutao Sun, Tianzhu Ye, Li~Dong, Hayden Kwok-Hay So, Yu~Hua, Ting Cao, Fan
  Yang, and Mao Yang.
\newblock Seerattention-r: Sparse attention adaptation for long reasoning.
\newblock \emph{arXiv preprint arXiv:2506.08889}, 2025.

\bibitem[Ge et~al.(2024{\natexlab{a}})Ge, Zhang, Liu, Zhang, Han, and
  Gao]{ge2024fastgen}
Suyu Ge, Yunan Zhang, Liyuan Liu, Minjia Zhang, Jiawei Han, and Jianfeng Gao.
\newblock Model tells you what to discard: Adaptive kv cache compression for
  llms.
\newblock In \emph{International Conference on Learning Representations
  (ICLR)}, 2024{\natexlab{a}}.

\bibitem[Ge et~al.(2024{\natexlab{b}})Ge, Hu, Wang, Wang, Chen, and
  Wei]{ge2024icae}
Tao Ge, Jing Hu, Lei Wang, Xun Wang, Si-Qing Chen, and Furu Wei.
\newblock In-context autoencoder for context compression in a large language
  model.
\newblock In \emph{International Conference on Learning Representations
  (ICLR)}, 2024{\natexlab{b}}.

\bibitem[Hao et~al.(2024)Hao, Sukhbaatar, Su, Li, Hu, Weston, and
  Tian]{hao2024coconut}
Shibo Hao, Sainbayar Sukhbaatar, DiJia Su, Xian Li, Zhiting Hu, Jason Weston,
  and Yuandong Tian.
\newblock Training large language models to reason in a continuous latent
  space.
\newblock \emph{arXiv preprint arXiv:2412.06769}, 2024.

\bibitem[Hooper et~al.(2024)Hooper, Kim, Mohammadzadeh, Mahoney, Shao, Keutzer,
  and Gholami]{hooper2024kvquant}
Coleman Hooper, Sehoon Kim, Hiva Mohammadzadeh, Michael~W. Mahoney,
  Yakun~Sophia Shao, Kurt Keutzer, and Amir Gholami.
\newblock Kvquant: Towards 10 million context length llm inference with kv
  cache quantization.
\newblock \emph{arXiv preprint arXiv:2401.18079}, 2024.
\newblock NeurIPS 2024.

\bibitem[Kim et~al.(2025)Kim, Kim, Kwon, Lee, Yun, and Song]{kim2025kvzip}
Jang-Hyun Kim, Jinuk Kim, Sangwoo Kwon, Jae~W. Lee, Sangdoo Yun, and Hyun~Oh
  Song.
\newblock Kvzip: Query-agnostic kv cache compression with context
  reconstruction.
\newblock In \emph{Advances in Neural Information Processing Systems}, 2025.

\bibitem[Kontonis et~al.(2026)Kontonis, Zeng, Garg, Chen, Tang, Wang,
  Awadallah, Horvitz, Langford, and Papailiopoulos]{kontonis2026memento}
Vasilis Kontonis, Yuchen Zeng, Shivam Garg, Lingjiao Chen, Hao Tang, Ziyan
  Wang, Ahmed Awadallah, Eric Horvitz, John Langford, and Dimitris
  Papailiopoulos.
\newblock {MEMENTO}: Teaching {LLMs} to manage their own context.
\newblock \emph{arXiv preprint arXiv:2604.09852}, 2026.

\bibitem[Korbak et~al.(2025)Korbak, Balesni, Barnes, Bengio, Benton, Bloom,
  et~al.]{korbak2025monitorability}
Tomek Korbak, Mikita Balesni, Elizabeth Barnes, Yoshua Bengio, Joe Benton,
  Joseph Bloom, et~al.
\newblock Chain of thought monitorability: A new and fragile opportunity for
  {AI} safety.
\newblock \emph{arXiv preprint arXiv:2507.11473}, 2025.

\bibitem[Kuzina et~al.(2026)Kuzina, Pioro, Whatmough, and
  Ehteshami~Bejnordi]{kuzina2026kava}
Anna Kuzina, Maciej Pioro, Paul~N. Whatmough, and Babak Ehteshami~Bejnordi.
\newblock Kava: Latent reasoning via compressed kv-cache distillation.
\newblock In \emph{International Conference on Learning Representations
  (ICLR)}, 2026.

\bibitem[Kwon et~al.(2023)Kwon, Li, Zhuang, Sheng, Zheng, Yu, Gonzalez, Zhang,
  and Stoica]{kwon2023vllm}
Woosuk Kwon, Zhuohan Li, Siyuan Zhuang, Ying Sheng, Lianmin Zheng, Cody~Hao Yu,
  Joseph~E. Gonzalez, Hao Zhang, and Ion Stoica.
\newblock Efficient memory management for large language model serving with
  {PagedAttention}.
\newblock In \emph{Proceedings of the 29th Symposium on Operating Systems
  Principles (SOSP)}, 2023.

\bibitem[Li et~al.(2024)Li, Huang, Yang, Venkitesh, Locatelli, Ye, Cai, Lewis,
  and Chen]{li2024snapkv}
Yuhong Li, Yingbing Huang, Bowen Yang, Bharat Venkitesh, Acyr Locatelli,
  Hanchen Ye, Tianle Cai, Patrick Lewis, and Deming Chen.
\newblock Snapkv: Llm knows what you are looking for before generation.
\newblock In \emph{Advances in Neural Information Processing Systems}, 2024.

\bibitem[Liu et~al.(2026)Liu, Zhao, Liu, Ruan, Huang, Dong, Xiao, Wang, Li,
  Zhu, and Xiao]{liu2026hybridthinker}
Xin Liu, Runsong Zhao, Xinyu Liu, Junhao Ruan, Pengcheng Huang, Shichao Dong,
  Chunyang Xiao, Chenglong Wang, Changliang Li, Jingbo Zhu, and Tong Xiao.
\newblock Hybridthinker: Efficient chain-of-thought reasoning via compressed
  memory and transient thought steps.
\newblock \emph{arXiv preprint arXiv:2606.03768}, 2026.

\bibitem[Liu et~al.(2023)Liu, Desai, Liao, Wang, Xie, Xu, Kyrillidis, and
  Shrivastava]{liu2023scissorhands}
Zichang Liu, Aditya Desai, Fangshuo Liao, Weitao Wang, Victor Xie, Zhaozhuo Xu,
  Anastasios Kyrillidis, and Anshumali Shrivastava.
\newblock Scissorhands: Exploiting the persistence of importance hypothesis for
  llm kv cache compression at test time.
\newblock \emph{Advances in Neural Information Processing Systems}, 36, 2023.

\bibitem[Liu et~al.(2024)Liu, Yuan, Jin, Zhong, Xu, Braverman, Chen, and
  Hu]{liu2024kivi}
Zirui Liu, Jiayi Yuan, Hongye Jin, Shaochen Zhong, Zhaozhuo Xu, Vladimir
  Braverman, Beidi Chen, and Xia Hu.
\newblock Kivi: A tuning-free asymmetric 2bit quantization for kv cache.
\newblock \emph{arXiv preprint arXiv:2402.02750}, 2024.
\newblock ICML 2024.

\bibitem[Mohtashami \& Jaggi(2023)Mohtashami and Jaggi]{mohtashami2023landmark}
Amirkeivan Mohtashami and Martin Jaggi.
\newblock Landmark attention: Random-access infinite context length for
  transformers.
\newblock In \emph{Advances in Neural Information Processing Systems}, 2023.

\bibitem[Monea et~al.(2025)Monea, Feldman, Padmanabhan, Brantley, and
  Artzi]{monea2025breadcrumbs}
Giovanni Monea, Yair Feldman, Shankar Padmanabhan, Kiant{\'e} Brantley, and
  Yoav Artzi.
\newblock Breadcrumbs reasoning: Memory-efficient reasoning with compression
  beacons.
\newblock \emph{arXiv preprint arXiv:2510.13797}, 2025.

\bibitem[Mu et~al.(2023)Mu, Li, and Goodman]{mu2023gist}
Jesse Mu, Xiang~Lisa Li, and Noah Goodman.
\newblock Learning to compress prompts with gist tokens.
\newblock In \emph{Advances in Neural Information Processing Systems}, 2023.

\bibitem[Muennighoff et~al.(2025)Muennighoff, Yang, Shi, Li, Fei-Fei,
  Hajishirzi, Zettlemoyer, Liang, Cand{\`e}s, and Hashimoto]{muennighoff2025s1}
Niklas Muennighoff, Zitong Yang, Weijia Shi, Xiang~Lisa Li, Li~Fei-Fei,
  Hannaneh Hajishirzi, Luke Zettlemoyer, Percy Liang, Emmanuel Cand{\`e}s, and
  Tatsunori Hashimoto.
\newblock s1: Simple test-time scaling.
\newblock In \emph{Proceedings of EMNLP}, 2025.

\bibitem[Nawrot et~al.(2024)Nawrot, {\L}a{\'n}cucki, Chochowski, Tarjan, and
  Ponti]{nawrot2024dmc}
Piotr Nawrot, Adrian {\L}a{\'n}cucki, Marcin Chochowski, David Tarjan, and
  Edoardo~M. Ponti.
\newblock Dynamic memory compression: Retrofitting llms for accelerated
  inference.
\newblock \emph{arXiv preprint arXiv:2403.09636}, 2024.
\newblock ICML 2024.

\bibitem[{OpenAI}(2024)]{openai2024o1}
{OpenAI}.
\newblock Learning to reason with {LLMs}.
\newblock \url{https://openai.com/index/learning-to-reason-with-llms/}, 2024.

\bibitem[Oren et~al.(2024)Oren, Hassid, Yarden, Adi, and
  Schwartz]{oren2024tova}
Matanel Oren, Michael Hassid, Nir Yarden, Yossi Adi, and Roy Schwartz.
\newblock Transformers are multi-state rnns.
\newblock In \emph{Proceedings of the 2024 Conference on Empirical Methods in
  Natural Language Processing}, pp.\  18724--18741, 2024.

\bibitem[Pang et~al.(2024)Pang, Ye, Wong, He, Chen, and Wang]{pang2024anllm}
Jianhui Pang, Fanghua Ye, Derek~F. Wong, Xin He, Wanshun Chen, and Longyue
  Wang.
\newblock Anchor-based large language models.
\newblock \emph{arXiv preprint arXiv:2402.07616}, 2024.
\newblock Findings of ACL 2024.

\bibitem[Qin et~al.(2024)Qin, Rosset, Chau, Rao, and Van~Durme]{qin2024dodo}
Guanghui Qin, Corby Rosset, Ethan~C. Chau, Nikhil Rao, and Benjamin Van~Durme.
\newblock Dodo: Dynamic contextual compression for decoder-only lms.
\newblock In \emph{Proceedings of ACL}, 2024.

\bibitem[Qin et~al.(2025)Qin, Cao, Lin, Hu, Fan, Cheng, Lin, and
  Li]{qin2025cake}
Ziran Qin, Yuchen Cao, Mingbao Lin, Wen Hu, Shixuan Fan, Ke~Cheng, Weiyao Lin,
  and Jianguo Li.
\newblock Cake: Cascading and adaptive kv cache eviction with layer
  preferences.
\newblock In \emph{International Conference on Learning Representations
  (ICLR)}, 2025.

\bibitem[Snell et~al.(2024)Snell, Lee, Xu, and Kumar]{snell2024scaling}
Charlie Snell, Jaehoon Lee, Kelvin Xu, and Aviral Kumar.
\newblock Scaling {LLM} test-time compute optimally can be more effective than
  scaling model parameters.
\newblock \emph{arXiv preprint arXiv:2408.03314}, 2024.

\bibitem[Song et~al.(2025)Song, Jo, Kim, and Kim]{song2025rpc}
Jiwon Song, Dongwon Jo, Yulhwa Kim, and Jae-Joon Kim.
\newblock Reasoning path compression: Compressing generation trajectories for
  efficient llm reasoning.
\newblock \emph{arXiv preprint arXiv:2505.13866}, 2025.
\newblock NeurIPS 2025.

\bibitem[Tan et~al.(2025)Tan, Li, Ju, Luo, Song, and Luan]{tan2025colar}
Wenhui Tan, Jiaze Li, Jianzhong Ju, Zhenbo Luo, Ruihua Song, and Jian Luan.
\newblock Think silently, think fast: Dynamic latent compression of llm
  reasoning chains.
\newblock In \emph{Advances in Neural Information Processing Systems}, 2025.

\bibitem[Tang et~al.(2025)Tang, Lin, Lin, Han, Hong, Yao, and
  Wang]{tang2025razorattention}
Hanlin Tang, Yang Lin, Jing Lin, Qingsen Han, Shikuan Hong, Yiwu Yao, and
  Gongyi Wang.
\newblock Razorattention: Efficient kv cache compression through retrieval
  heads.
\newblock In \emph{International Conference on Learning Representations
  (ICLR)}, 2025.

\bibitem[Tang et~al.(2024)Tang, Zhao, Zhu, Xiao, Kasikci, and
  Han]{tang2024quest}
Jiaming Tang, Yilong Zhao, Kan Zhu, Guangxuan Xiao, Baris Kasikci, and Song
  Han.
\newblock Quest: Query-aware sparsity for efficient long-context llm inference.
\newblock In \emph{International Conference on Machine Learning (ICML)}, 2024.

\bibitem[Wan et~al.(2025)Wan, Wu, Zhang, Xin, Tao, Zhu, Wang, Luo, Xiong, Wang,
  and Zhang]{wan2025d2o}
Zhongwei Wan, Xinjian Wu, Yu~Zhang, Yi~Xin, Chaofan Tao, Zhihong Zhu, Xin Wang,
  Siqi Luo, Jing Xiong, Longyue Wang, and Mi~Zhang.
\newblock D2o: Dynamic discriminative operations for efficient long-context
  inference of large language models.
\newblock In \emph{International Conference on Learning Representations
  (ICLR)}, 2025.

\bibitem[Wang et~al.(2026)Wang, Qiu, Zhao, Qian, Yang, Han, Ji, Savarese,
  Heinecke, and Wang]{wang2026randomattention}
Heng Wang, Jielin Qiu, Wenting Zhao, Cheng Qian, Liangwei Yang, Jiawei Han,
  Heng Ji, Silvio Savarese, Shelby Heinecke, and Huan Wang.
\newblock Random attention: Rethinking kv cache eviction for efficient
  reasoning.
\newblock \emph{arXiv preprint arXiv:2609.03430}, 2026.

\bibitem[Wang et~al.(2024)Wang, Cui, and Gan]{wang2024squeezeattention}
Zihao Wang, Bin Cui, and Shaoduo Gan.
\newblock Squeezeattention: 2d management of kv-cache in llm inference via
  layer-wise optimal budget.
\newblock \emph{arXiv preprint arXiv:2404.04793}, 2024.

\bibitem[Xia et~al.(2025)Xia, Leong, Wang, Li, and Li]{xia2025tokenskip}
Heming Xia, Chak~Tou Leong, Wenjie Wang, Yongqi Li, and Wenjie Li.
\newblock Tokenskip: Controllable chain-of-thought compression in llms.
\newblock In \emph{Proceedings of EMNLP}, 2025.

\bibitem[Xiao et~al.(2024{\natexlab{a}})Xiao, Zhang, Han, Xiao, Lin, Zhang,
  Liu, and Sun]{xiao2024infllm}
Chaojun Xiao, Pengle Zhang, Xu~Han, Guangxuan Xiao, Yankai Lin, Zhengyan Zhang,
  Zhiyuan Liu, and Maosong Sun.
\newblock Infllm: Training-free long-context extrapolation for llms with an
  efficient context memory.
\newblock In \emph{Advances in Neural Information Processing Systems},
  2024{\natexlab{a}}.

\bibitem[Xiao et~al.(2024{\natexlab{b}})Xiao, Tian, Chen, Han, and
  Lewis]{xiao2024streamingllm}
Guangxuan Xiao, Yuandong Tian, Beidi Chen, Song Han, and Mike Lewis.
\newblock Efficient streaming language models with attention sinks.
\newblock In \emph{International Conference on Learning Representations
  (ICLR)}, 2024{\natexlab{b}}.

\bibitem[Xu et~al.(2025)Xu, Xie, Zhao, and He]{xu2025cod}
Silei Xu, Wenhao Xie, Lingxiao Zhao, and Pengcheng He.
\newblock Chain of draft: Thinking faster by writing less.
\newblock \emph{arXiv preprint arXiv:2502.18600}, 2025.

\bibitem[Yan et~al.(2025)Yan, Shen, Liu, Jiang, Zhang, Shao, and
  Zhuang]{yan2025inftythink}
Yuchen Yan, Yongliang Shen, Yang Liu, Jin Jiang, Mengdi Zhang, Jian Shao, and
  Yueting Zhuang.
\newblock {InftyThink}: Breaking the length limits of long-context reasoning in
  large language models.
\newblock \emph{arXiv preprint arXiv:2503.06692}, 2025.

\bibitem[Yan et~al.(2026)Yan, Jiang, Jiang, Li, Wen, Zhang, Zhou, Shao, Zhuang,
  and Shen]{yan2026inftythinkplus}
Yuchen Yan, Liang Jiang, Jin Jiang, Shuaicheng Li, Zujie Wen, Zhiqiang Zhang,
  Jun Zhou, Jian Shao, Yueting Zhuang, and Yongliang Shen.
\newblock {InftyThink+}: Effective and efficient infinite-horizon reasoning via
  reinforcement learning.
\newblock In \emph{International Conference on Machine Learning (ICML)}, 2026.

\bibitem[Yang et~al.(2024)Yang, Han, Gao, Hu, Zhang, and
  Zhao]{yang2024pyramidinfer}
Dongjie Yang, Xiaodong Han, Yan Gao, Yao Hu, Shilin Zhang, and Hai Zhao.
\newblock Pyramidinfer: Pyramid kv cache compression for high-throughput llm
  inference.
\newblock In \emph{Findings of ACL}, 2024.

\bibitem[Yang et~al.(2026)Yang, Guo, Huang, Wang, Shi, Wang, Liang, and
  Tang]{yang2026accordion}
Zhicheng Yang, Zhijiang Guo, Yinya Huang, Yongxin Wang, Wenlei Shi, Yiwei Wang,
  Xiaodan Liang, and Jing Tang.
\newblock {Accordion-Thinking}: Self-regulated step summaries for efficient and
  readable {LLM} reasoning.
\newblock \emph{arXiv preprint arXiv:2602.03249}, 2026.

\bibitem[Ye et~al.(2025)Ye, Gan, Huang, Ge, and Tang]{ye2024vocollama}
Xubing Ye, Yukang Gan, Xiaoke Huang, Yixiao Ge, and Yansong Tang.
\newblock Voco-llama: Towards vision compression with large language models.
\newblock In \emph{Proceedings of the IEEE/CVF Conference on Computer Vision
  and Pattern Recognition (CVPR)}, pp.\  29836--29846, 2025.

\bibitem[Yuan et~al.(2025)Yuan, Gao, Dai, Luo, Zhao, Zhang, Xie, Wei, Wang,
  Xiao, Wang, Ruan, Zhang, Liang, and Zeng]{yuan2025nsa}
Jingyang Yuan, Huazuo Gao, Damai Dai, Junyu Luo, Liang Zhao, Zhengyan Zhang,
  Zhenda Xie, Y.~X. Wei, Lean Wang, Zhiping Xiao, Yuqing Wang, Chong Ruan, Ming
  Zhang, Wenfeng Liang, and Wangding Zeng.
\newblock Native sparse attention: Hardware-aligned and natively trainable
  sparse attention.
\newblock \emph{arXiv preprint arXiv:2502.11089}, 2025.
\newblock ACL 2025, Best Paper Award.

\bibitem[Zhang et~al.(2026)Zhang, Zhang, Ma, Zhang, and
  Guo]{zhang2026lazyeviction}
Haoyue Zhang, Hualei Zhang, Xiaosong Ma, Jie Zhang, and Song Guo.
\newblock Lazyeviction: Lagged kv eviction with attention pattern observation
  for efficient long reasoning.
\newblock In \emph{Proceedings of ACL}, 2026.

\bibitem[Zhang et~al.(2025)Zhang, Zhu, Sun, Luo, Qiao, Du, Zheng, Chen, and
  Zhang]{zhang2025lightthinker}
Jintian Zhang, Yuqi Zhu, Mengshu Sun, Yujie Luo, Shuofei Qiao, Lun Du,
  Da~Zheng, Huajun Chen, and Ningyu Zhang.
\newblock Lightthinker: Thinking step-by-step compression.
\newblock In \emph{Proceedings of EMNLP}, 2025.

\bibitem[Zhang et~al.(2024)Zhang, Liu, Xiao, Shao, Ye, and
  Dou]{zhang2024activationbeacon}
Peitian Zhang, Zheng Liu, Shitao Xiao, Ninglu Shao, Qiwei Ye, and Zhicheng Dou.
\newblock Long context compression with activation beacon.
\newblock \emph{arXiv preprint arXiv:2401.03462}, 2024.
\newblock ICLR 2025.

\bibitem[Zhang et~al.(2023)Zhang, Sheng, Zhou, Chen, Zheng, Cai, Song, Tian,
  R{\'e}, Barrett, Wang, and Chen]{zhang2023h2o}
Zhenyu Zhang, Ying Sheng, Tianyi Zhou, Tianlong Chen, Lianmin Zheng, Ruisi Cai,
  Zhao Song, Yuandong Tian, Christopher R{\'e}, Clark Barrett, Zhangyang Wang,
  and Beidi Chen.
\newblock H$_2$o: Heavy-hitter oracle for efficient generative inference of
  large language models.
\newblock \emph{Advances in Neural Information Processing Systems}, 36, 2023.

\end{thebibliography}
\bibliographystyle{iclr2027_conference}

\appendix
\section{Related work}
\label{sec:related}
\label{app:related}
We sort prior work by what it compresses and whether the compression is learned. Every method that learns a compressed
form of the model's own output was validated on generations one to two orders of magnitude shorter than ours, and every method
validated at our length selects among existing rows rather than learning new ones.

\paragraph{Learned compression of the model's own output.} MEMENTO~\citep{kontonis2026memento} trains 8B--32B models to
segment their reasoning into blocks, write a textual summary of each block and continue from the summaries alone, with about
$2.5\times$ lower peak KV; like the other text-summary methods of the introduction, its summary is text the model writes, whereas
ours is a set of learned KV rows. Its models lose accuracy against their full-context counterparts on competition mathematics
before reinforcement learning, whereas our recipe matches its full-KV counterpart with supervised training alone.
LightThinker~\citep{zhang2025lightthinker} is the nearest learned-row method: at each thought boundary the model writes the
thought into 9 cache tokens and drops the thought's raw KV, with outputs of at most 10K tokens. HybridThinker~\citep{liu2026hybridthinker}
keeps the last few steps raw, at about 2.5K-token outputs, and Breadcrumbs Reasoning~\citep{monea2025breadcrumbs} keeps one
beacon token every 2--32 generated tokens, on synthetic tasks with at most 1{,}000 generated tokens. Dynamic Memory
Compression~\citep{nawrot2024dmc} learns per head and step whether to append a KV pair or merge it into the last slot; it merges
raw rows rather than writing new ones. Latent-reasoning methods replace the chain itself with
vectors~\citep{hao2024coconut,cheng2024ccot,tan2025colar,kuzina2026kava}, give up the readable trace, and are evaluated at
GSM8K scale. None of these reports accuracy against a full-KV model on generations of our length.

\paragraph{Compression tokens for inputs, and where their rows sit.} Our code rows are an instance of a device introduced for
inputs: place a few learned tokens after a span, hide the span from everything that follows, and train end to end so its
information must route through those tokens. Gist tokens~\citep{mu2023gist}, ICAE~\citep{ge2024icae},
AutoCompressors~\citep{chevalier2023autocompressors}, Activation Beacon~\citep{zhang2024activationbeacon} and
VoCo-LLaMA~\citep{ye2024vocollama} all compress a fixed input once, before generation, and are evaluated with short outputs; we
compress the generated stream tens of times inside one response. None of them places its rows inside the span they replace:
gist tokens are ordinary tokens, accumulated Activation Beacons occupy consecutive positions in the compacted stream, and
LightThinker writes its rows at the thought boundary. The closest methods reuse an existing token's
position~\citep{pang2024anllm,qin2024dodo} or defer position to retrieval time~\citep{mohtashami2023landmark}. Our appended-position
ablation (Section~\ref{sec:ablations}) is therefore our own model with this prior convention swapped in, not a reproduction of any
one method. The nearest published configuration, LightThinker's token-level variant (every 6 generated tokens into 2 cache
tokens at appended positions), is reported by its authors 5--6 points below their thought-level scheme on GSM8K-class tasks.\label{app:positions}

\paragraph{Eviction and prompt compression.} Generic evictors keep recent tokens and heavy hitters by accumulated
attention~\citep{zhang2023h2o,liu2023scissorhands,oren2024tova,xiao2024streamingllm,ge2024fastgen,adnan2024keyformer}, and
reasoning-specific methods score tokens by importance and redundancy~\citep{cai2025rkv}, compress from a window of recent
queries~\citep{song2025rpc}, or evict on Qwen3-4B with outputs of up to 16K tokens~\citep{zhang2026lazyeviction,chang2026vase};
random eviction within the reasoning phase has been found to match the scored methods~\citep{wang2026randomattention}. Prompt
compressors shrink a long input once after prefill~\citep{li2024snapkv,cai2024pyramidkv,feng2025adakv,fu2025headkv,kim2025kvzip}
and do not touch the generated stream. All of these choose among rows that exist; our model writes new rows, trained for the
queries that will read them.

\paragraph{Allocating the budget across layers and heads.} Heterogeneous budgets have been set per head from attention
statistics~\citep{ge2024fastgen,feng2025adakv} or retrieval-head probes~\citep{fu2025headkv,tang2025razorattention}, and per layer
from fixed pyramids~\citep{cai2024pyramidkv,yang2024pyramidinfer} or measured
statistics~\citep{wang2024squeezeattention,wan2025d2o,qin2025cake}. Reported gains over uniform are modest outside very small
budgets, and no decode-time reasoning compressor ablates the allocation. We measure each layer's value on the trained model
itself, by cutting its rows (Section~\ref{sec:quotas}).

\paragraph{Complementary reductions.} Sparse-attention methods keep every row and reduce what each step
reads~\citep{tang2024quest,xiao2024infllm,yuan2025nsa,gao2025seerattentionr}; quantization~\citep{liu2024kivi,hooper2024kvquant},
latent KV~\citep{deepseek2024v2} and sharing across heads or layers~\citep{ainslie2023gqa,brandon2024cla} shrink each row; and
shorter chains of thought~\citep{xia2025tokenskip,xu2025cod} reduce the number of generated tokens. Each composes with a
reduction in the number of rows kept.

\FloatBarrier
\section{Experimental setup}
\FloatBarrier
\subsection{Training}
\label{app:train}
\begin{table}[h]
\centering\small
\caption{Training and evaluation setup shared by every arm.}
\label{tab:setup}
\begin{tabular}{@{}l >{\raggedright\arraybackslash}p{0.55\linewidth}@{}}
\toprule
Base model & Qwen3-4B-Base \\
SFT data & OpenMathReasoning-derived math SFT, 400K prompt/response pairs, fixed order \\
Response tokens per arm & 2B (about 936 steps at effective batch 256) \\
Block size $C$ / code slots $K_{\max}$ / mean quota & 512 / 256 / 128 \\
Optimizer, schedule & AdamW, lr 1e-4 WSD \\
Generation cap at eval & 32768 tokens \\
Hardware & one node per run, either 4$\times$ NVIDIA GB300 or 8$\times$ NVIDIA H100-80GB;
FlexAttention, bf16 autocast. A fixed data order sets each step's examples independently of the number of GPUs, so an
arm is the same run on either machine \\
\bottomrule
\end{tabular}
\end{table}

The learning rate warms up linearly over the first 3\% of the 2B response tokens, holds at $10^{-4}$ until 80\% of the budget,
and decays linearly to $10^{-5}$ over the last 20\%; AdamW with $\beta=(0.9, 0.95)$, weight decay 0.1 and gradient clipping at
1.0; effective batch 256 sequences (about 2.1M response tokens per step), 936 steps. The held-out set is the first 192 pairs.

\paragraph{Seed-to-seed noise.}\label{app:archived} The data seed sets the random stream that draws each training step's
examples from the 400K-pair pool, so two seeds train on the same pool and the same token budget but on different draws in a
different order. Retraining the full-KV model on a different data seed shifts the suite score by $0.8$ points (1 s.d.\ of the
paired difference), the grey band of the figures. Two additions to the loss were tried under the same protocol and did not
help: keeping the last 16 tokens of each block uncompressed (41.6 against 41.5 for the uniform setting), and distillation from
the frozen full-KV model (39.9).

\FloatBarrier
\subsection{Competition suite and protocol}
\label{app:suite}
\paragraph{Construction.} The set list was fixed before any arm was evaluated on it: AIME 2022--2026, HMMT February 2023--2025,
BRUMO, CMIMC and SMT 2025 from MathArena, and AMC 12 2022--2023 from the AI-MO validation sets (Table~\ref{tab:suite_desc}).
Two further sets, APEX 2025 and BeyondAIME, were left out before evaluation because a 4B model fine-tuned on our data solves too
few of their problems to separate arms. Every arm is scored on the same problems with the same samples, cap and grader, so every
comparison in the paper is a paired difference on identical problems.

\begin{table}[h]
\centering\footnotesize\setlength{\tabcolsep}{3.5pt}
\caption{The 446-problem competition suite by competition: source, release period, number of problems, problems that also
occur in the SFT prompts (word 13-gram overlap $\ge 50\%$ with a single training prompt), and avg@8 at the 32K cap for each arm.
Arms: full KV (vanilla SFT), \method{} per-layer, uniform allocation, appended code positions, LightThinker (thought-level, our
stack) and R-KV at budget 4096 on the vanilla weights. A single set's 95\% interval on a paired gap is about $\pm 5$ points
(Appendix~\ref{app:pooled}), so per-set differences are not individually meaningful.}
\label{tab:suite_desc}
\resizebox{\linewidth}{!}{\begin{tabular}{@{}lllrrrrrrrr@{}}
\toprule
Set & source & released & problems & in SFT & full KV & per-layer & uniform & appended & LightThinker & R-KV 4096 \\
\midrule
AIME 2022 & AI-MO/aimo-validation-aime (url 2022) & pre-2025 & 30 & 19 & 42.1 & 41.7 & 42.1 & 33.8 & 25.8 & 40.4 \\
AIME 2023 & MathArena/aime\_2023\_I+II & pre-2025 & 30 & 17 & 40.0 & 40.4 & 35.8 & 31.7 & 21.7 & 36.2 \\
AIME 2024 & MathArena/aime\_2024\_I+II & pre-2025 & 30 & -- & 50.8 & 50.0 & 42.9 & 34.2 & 36.7 & 52.9 \\
AIME 2025 & MathArena/aime\_2025 & 2025+ & 30 & -- & 37.1 & 33.8 & 34.2 & 27.5 & 25.0 & 30.8 \\
AIME 2026 & MathArena/aime\_2026 & 2025+ & 30 & -- & 36.2 & 38.3 & 37.1 & 24.2 & 29.2 & 33.8 \\
AMC 12 2022--2023 & AI-MO/aimo-validation-amc & pre-2025 & 83 & 32 & 75.6 & 75.2 & 73.2 & 67.5 & 66.1 & 76.7 \\
HMMT Feb 2023 & MathArena/hmmt\_feb\_2023 & pre-2025 & 30 & -- & 32.5 & 32.1 & 29.2 & 22.9 & 25.4 & 32.9 \\
HMMT Feb 2024 & MathArena/hmmt\_feb\_2024 & pre-2025 & 30 & -- & 27.1 & 24.6 & 21.7 & 20.4 & 18.8 & 26.2 \\
HMMT Feb 2025 & MathArena/hmmt\_feb\_2025 & 2025+ & 30 & -- & 22.5 & 22.9 & 24.6 & 15.4 & 16.2 & 22.1 \\
BRUMO 2025 & MathArena/brumo\_2025 & 2025+ & 30 & 1 & 50.4 & 45.0 & 44.2 & 41.7 & 36.7 & 44.2 \\
CMIMC 2025 & MathArena/cmimc\_2025 & 2025+ & 40 & -- & 19.1 & 21.6 & 20.9 & 16.2 & 13.8 & 18.8 \\
SMT 2025 & MathArena/smt\_2025 & 2025+ & 53 & -- & 44.1 & 42.2 & 42.2 & 36.8 & 32.8 & 43.4 \\
\bottomrule
\end{tabular}}
\end{table}

\paragraph{Why pool.}\label{app:pooled} On our paired data a single 30-problem set puts the 95\% interval on a gap between two
arms at roughly $\pm 5$ points: a full-KV replicate differing only in its data seed comes out significantly worse than the
original on AIME 2024 and better on AIME 2025. Pooling 446 problems brings the half-width to $1.5$ points. Cheaper metrics do not
substitute for it: across a $4\times$ range of budget, held-out CE moves by $0.010$ while the suite moves by $4.0$ points
(Appendix~\ref{app:scaling}).

\paragraph{Table~\ref{tab:main} protocol.} All SFT arms fine-tune Qwen3-4B-Base on the same 2B response tokens in the same
order, with $C{=}512$. KV/512 counts persistent KV rows per 512 generated tokens per KV head; for R-KV, which keeps a constant
number of rows, it is the measured residency over its generations. MATH-500 is avg@4 with the strict last-boxed grader. The three
simple eviction policies were run in the HuggingFace decoder on a 60-problem subset (AIME 2025 and HMMT Feb 2025), paired against
full KV on those problems ($^\dagger$ in the table).

\paragraph{Overlap with the training data.} A problem is flagged when at least half of its word 13-grams occur in a single
SFT training prompt. 69 of the 446 problems are flagged, all from AIME 2022--2023 and AMC except one borderline BRUMO problem;
none from AIME 2024--2026, HMMT, CMIMC or SMT. Overlap raises absolute accuracy for every arm (full KV 69.2 on flagged problems,
39.2 on the rest), but all arms share the training data, so paired differences remain like for like. Table~\ref{tab:suite_subsets}
repeats the analysis on subsets. Every difference to full KV keeps its sign on every subset, and the significant ones stay
significant without the overlapping problems. The per-layer-versus-uniform difference is positive on every subset ($+1.6$
overall, $+1.1$ without overlap, $+2.9$ before 2025, $+0.1$ on 2025+ problems); each subset holds about half the suite, so its
interval spans about four points and only the pooled suite resolves a difference of this size.

\begin{table}[h]
\centering\footnotesize\setlength{\tabcolsep}{3pt}
\caption{Paired differences on subsets of the suite (avg@8 points, 95\% bootstrap interval over problems). The first five
columns are differences to full KV on the same problems; the last is per-layer minus uniform. ``Overlapping'' problems are the 69
that occur in the SFT prompts (Table~\ref{tab:suite_desc}).}
\label{tab:suite_subsets}
\resizebox{\linewidth}{!}{\begin{tabular}{@{}lrrrrrrrr@{}}
\toprule
& & & \multicolumn{5}{c}{difference to full KV} & \\
\cmidrule(lr){4-8}
Subset & problems & full KV & per-layer & uniform & appended & LightThinker & R-KV 4096 & per-layer $-$ uniform \\
\midrule
All 446 problems & 446 & 43.8 & $-0.8$ {\scriptsize $[-2.3, +0.8]$} & $-2.3$ {\scriptsize $[-3.8, -0.9]$} & $-8.5$ {\scriptsize $[-10.3, -6.8]$} & $-10.5$ {\scriptsize $[-12.5, -8.7]$} & $-1.2$ {\scriptsize $[-2.7, +0.3]$} & $+1.6$ {\scriptsize $[+0.1, +3.0]$} \\
No overlap with the SFT data & 377 & 39.2 & $-0.8$ {\scriptsize $[-2.4, +0.9]$} & $-1.9$ {\scriptsize $[-3.5, -0.3]$} & $-7.8$ {\scriptsize $[-9.6, -5.9]$} & $-9.2$ {\scriptsize $[-11.3, -7.2]$} & $-1.0$ {\scriptsize $[-2.6, +0.6]$} & $+1.1$ {\scriptsize $[-0.4, +2.8]$} \\
Released 2025 or later & 213 & 35.2 & $-0.9$ {\scriptsize $[-3.3, +1.3]$} & $-1.0$ {\scriptsize $[-3.2, +1.1]$} & $-7.6$ {\scriptsize $[-10.2, -5.2]$} & $-9.3$ {\scriptsize $[-12.3, -6.6]$} & $-2.4$ {\scriptsize $[-4.8, -0.3]$} & $+0.1$ {\scriptsize $[-1.9, +2.3]$} \\
Released before 2025 & 233 & 51.7 & $-0.6$ {\scriptsize $[-2.7, +1.3]$} & $-3.5$ {\scriptsize $[-5.5, -1.5]$} & $-9.3$ {\scriptsize $[-11.7, -6.9]$} & $-11.6$ {\scriptsize $[-14.2, -8.9]$} & $-0.1$ {\scriptsize $[-2.1, +2.0]$} & $+2.9$ {\scriptsize $[+0.9, +5.0]$} \\
Overlapping problems only & 69 & 69.2 & $-0.7$ {\scriptsize $[-4.5, +2.9]$} & $-4.7$ {\scriptsize $[-8.2, -1.3]$} & $-12.5$ {\scriptsize $[-17.2, -7.8]$} & $-18.1$ {\scriptsize $[-23.6, -13.0]$} & $-2.5$ {\scriptsize $[-6.7, +1.4]$} & $+4.0$ {\scriptsize $[+0.5, +7.4]$} \\
\bottomrule
\end{tabular}}
\end{table}

\FloatBarrier
\section{Baselines}
\FloatBarrier
\subsection{R-KV}
\label{app:rkv}
R-KV~\citep{cai2025rkv} scores each generated token by attention importance minus key redundancy and keeps the highest-scoring
rows up to a fixed budget. We port it to the vLLM version we serve with and run it on the full-KV checkpoint with the same
problems, sampling and grader (Table~\ref{tab:rkvcost}).

\begin{table}[t]
\centering\footnotesize
\caption{R-KV against the arm it is paired with at each budget, on the same 446 problems, the same sampling and the same
grader. KV/512 = persistent rows per 512 generated tokens per KV head; R-KV's is its \emph{measured} mean residency, so at
both settings it holds about 12\% more cache than the arm beside it. trunc.\ = \% of samples that reach the 32K cap.
$\Delta$ is the paired per-problem difference against that partner arm, bootstrapped on the 446 common problems.
Accuracy separates them only at the tighter budget; truncation and generated length separate them at both.}
\label{tab:rkvcost}
\begin{tabular}{@{}lrlcrr@{}}
\toprule
Arm & KV/512 & Suite avg@8 & $\Delta$ vs.\ partner & trunc.\ \% & gen.\ tokens \\
\midrule
vanilla SFT (full KV) & 512 & 43.8 {\scriptsize[39.8, 47.6]} & -- & 3.4 & 13,162 \\
\method{}, per-layer 4:1 & 128 & 43.0 {\scriptsize[39.1, 46.9]} & -- & 3.2 & 13,257 \\
R-KV, budget 4096 & 144 & 42.6 {\scriptsize[38.5, 46.6]} & $-0.4$ {\scriptsize[-2.1, +1.2]} & 31.7 & 18,428 \\
\method{}, uniform 8:1 & 64 & 39.4 {\scriptsize[35.5, 43.2]} & -- & 3.6 & 13,673 \\
R-KV, budget 2048 & 74 & 35.6 {\scriptsize[31.7, 39.5]} & $-3.8$ {\scriptsize[-5.5, -2.2]} & 55.8 & 22,230 \\
\bottomrule
\end{tabular}
\end{table}

\paragraph{Budget matching.} R-KV's budget is a constant number of rows; ours is a rate. We match them by measured residency:
over the suite's samples a budget of 4096 leaves about 12\% more generated rows resident than our 4:1 arm, and 2048 about 12\%
more than our 8:1 arm, so each R-KV setting has slightly more cache than the arm it is compared with.

\paragraph{Served form.} R-KV's reference code selects tokens per KV head, which a paged cache cannot honour, since dropping a
token frees it for every head; we follow R-KV's own SGLang and vLLM ports, which reduce the score to one decision per token. We
enable its opt-in flag that returns evicted pages to the allocator, without which eviction saves no memory. Our port reproduces
the released scoring bit for bit on every configuration we tested, uses the hyperparameters of R-KV's own evaluation, and
compresses every 64 decoding steps rather than 128, which keeps the cache closer to the budget at a higher cost per token. Every
request in both reported runs was compacted.

\FloatBarrier
\subsection{LightThinker}
\label{app:lightthinker}
LightThinker~\citep{zhang2025lightthinker} compresses a model's own reasoning online into learned KV entries. We compare against
its mechanism trained on our data and served in our engine rather than against its released checkpoints, which differ in base
model, data and decoding.

\paragraph{The arm.} A thought ends at a paragraph break; the model then writes 9 cache tokens at appended positions under
LightThinker's attention mask and continues from them, and the thought's raw KV is dropped at inference. Data, recipe and the
2B-token budget are those of every other arm; the served model matches the training-time forward pass (argmax agreement 0.996).
Thoughts end every 53 tokens on average, so the arm keeps 97 rows per 512 generated tokens, between our 8:1 and 4:1 arms.

\paragraph{Result.} It scores $33.3$ on the suite, $-9.8$ $[-11.5, -8.0]$ paired against \method{}, and $2.0$ points below our
appended-position ablation, which shares its positions but compresses 512-token blocks. The compression unit explains most of the
gap: at a fixed ratio, the block size sets how many lossy compressions a late query reads through and how much recent text stays
exact (Table~\ref{tab:blockscale}). It also writes longer, 14.9K tokens on average against 13.2K for full KV
(Figure~\ref{fig:lenflow}a). Its time to finish is an upper bound: its compression is triggered by a sampled token, so it ran
with synchronous scheduling while every other arm ran asynchronously.

\begin{table}[h]
\centering\small
\caption{Compression events a late query reads through, for our mean generation (13.2K tokens) and at the 32K cap, and the
exact window (the open block, $C/2$ tokens in expectation).}
\label{tab:blockscale}
\begin{tabular}{@{}lrrr@{}}
\toprule
block $C$ & events (mean gen.) & events (at cap) & exact window \\
\midrule
1024 & 13 & 31 & 512 \\
512 (ours) & 26 & 62 & 256 \\
256 & 51 & 124 & 128 \\
$\approx$53 (LightThinker, measured) & 249 & 618 & 26 \\
\bottomrule
\end{tabular}
\end{table}

\FloatBarrier
\section{Additional results}
\FloatBarrier
\subsection{Per-layer versus per-head allocation}
\label{app:dynamic}
\begin{table}[t]
\centering\small
\caption{Budget allocation at a fixed mean of 128 code rows per block (all other settings identical). Held-out CE on 192 held-out SFT pairs;
MATH-500 avg@4; suite avg@8 at 32K as in Table~\ref{tab:main}. The per-head row lets a head keep up to all 512 rows, with a floor of 32,
nested inside the per-layer schedule; paired against the per-layer headline it is $-0.6$ $[-2.1, +0.9]$. The last row sets per-head quotas from how
concentrated each head's attention is in the full-KV model, which raises the early layers; paired against the per-layer model it is $-1.9$ $[-3.3, -0.5]$.}
\label{tab:allocation}
\begin{tabular}{@{}lrrl@{}}
\toprule
Schedule & held-out CE & MATH-500 & Suite avg@8 [95\% CI] \\
\midrule
uniform (every head 128) & 0.4227 & 88.3 & 41.5 {\scriptsize[37.6, 45.3]} \\
per layer, measured on the trained model (\method{}) & 0.4215 & 87.5 & 43.0 {\scriptsize[39.1, 46.9]} \\
per head, measured, $K_{\max}{=}512$ (nested) & 0.4209 & -- & 42.4 {\scriptsize[38.5, 46.3]} \\
per head, from full-KV attention statistics & 0.4231 & -- & 41.1 {\scriptsize[37.2, 45.0]} \\
\bottomrule
\end{tabular}
\end{table}

\begin{figure}[h]
  \centering
  \includegraphics[width=0.9\linewidth]{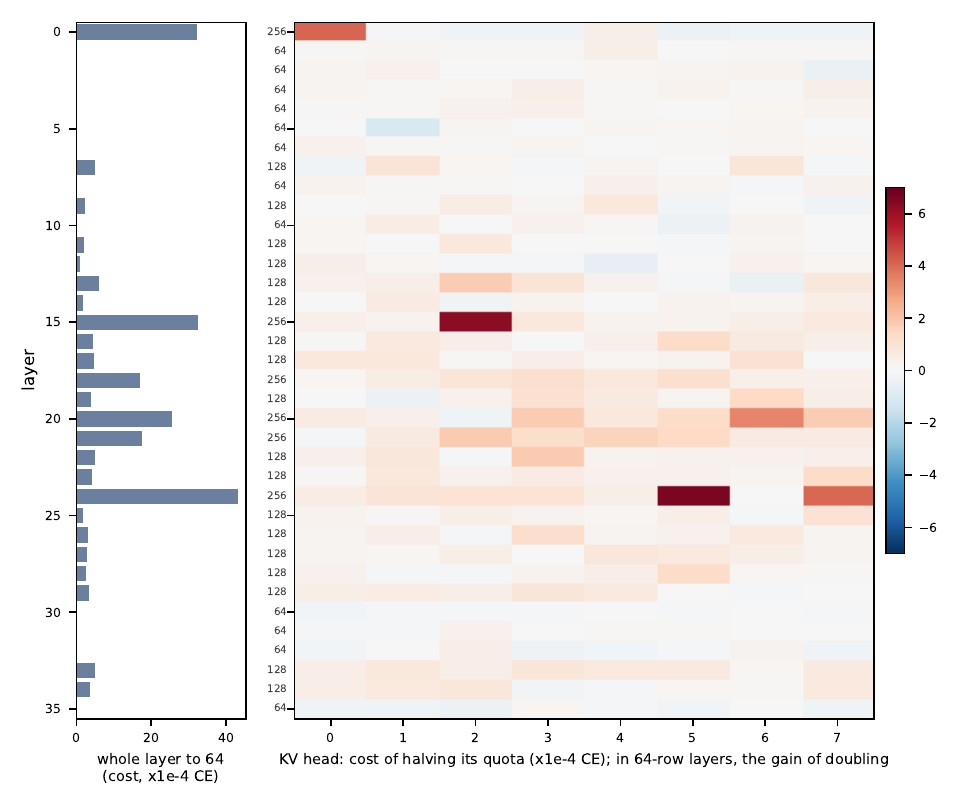}
  \caption{Where a layer's code rows are worth something, head by head. On the trained per-layer model, each (layer, KV head)
  cell shows the held-out CE cost of halving that one head's quota (in 64-row layers, the gain of doubling it), on the full
  192-pair held-out set; the bars at left repeat the whole-layer cost of Figure~\ref{fig:budget}. Noise floor: $0.4\times10^{-4}$
  (s.d.\ of head cells in layers whose whole-layer cost is nil).}
  \label{fig:perhead}
\end{figure}
Figure~\ref{fig:perhead} repeats the truncation measurement of Section~\ref{sec:allocation} head by head on the per-layer model.
In the layers cut to 64 rows, doubling any single head gains nothing; in the layers at 128, no head is expensive; in the six
layers raised to 256, the value sits in one or two heads. So the per-layer schedule already captures nearly all of the value
that can be moved at this mean, and a per-head schedule trained from scratch (Table~\ref{tab:allocation}) is within noise of it.
It would also cost to serve: a paged engine keeps one row count per layer, so ragged per-head quotas either pad every head to the
layer's largest ($1.53\times$ the memory of the per-layer schedule) or need a packed side cache, which adds about 29~ms of fixed
overhead per decode step. We therefore allocate per layer.

\FloatBarrier
\subsection{What the code rows carry}
\label{app:probe}
\begin{figure}[h]
  \centering
  \includegraphics[width=\linewidth]{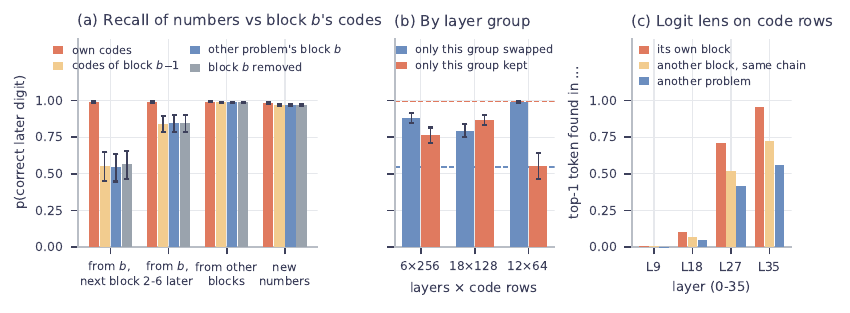}
  \caption{What the code rows carry (115 AIME chains, 895 intervened blocks). (a)~Probability of a number's later digits, by
  where it was first written, with block $b$'s code rows intact, replaced, or removed (95\% intervals over chains). (b)~Numbers
  from $b$ reused in the next block, with one layer group swapped or only one kept; dashed: all intact, all swapped. (c)~How
  often a code row's logit-lens top token occurs in its own block, another block, or another problem.}
  \label{fig:probe}
\end{figure}
Figure~\ref{fig:probe} adds two views to Section~\ref{sec:probe}. Swapping block $b$'s code rows in only the 12 layers cut to
64 rows leaves next-block recall at $0.99$, and keeping only those layers intact is as bad as swapping every layer ($0.55$); the
256- and 128-row layers carry the content (kept alone, $0.77$ and $0.87$). Projected onto the vocabulary after layer 27, a code
row's most likely token occurs in its own block for 71\% of blocks, against 52\% for another block of the same chain and 42\%
for another problem's.

\FloatBarrier
\subsection{Block size, code budget and training cost}
\label{app:scaling}
Within one block size, halving the code budget costs far more on the suite than held-out CE suggests. At $C{=}512$, going
from $K{=}128$ (4:1) to $K{=}64$ (8:1) moves held-out CE by $0.005$ but the pooled suite by $2.1$ points, $41.5$ to $39.4$.
Doubling the budget the other way, $K{=}256$ (2:1) moves held-out CE by a further $0.004$ and the suite by $1.9$ points to
$43.4$, which is $-0.4$ $[-1.8, +1.0]$ against full KV: lossless within the interval, at half the compression. The remaining
rows of Table~\ref{tab:scaling} vary the block size at a fixed ratio (Figure~\ref{fig:blockscale}); held-out CE is not
comparable across block sizes, since the held-out set is defined by blocks per example.
\begin{table}[t]
\centering\small
\caption{Block size and code budget sweep (uniform quotas, bitrev positions, $K_{\max}{=}256$, same 2B tokens). Held-out CE is comparable only
within one block size (the held-out set is defined by blocks per example). MATH-500 avg@4; suite avg@8 at 32K as in Table~\ref{tab:main}.}
\label{tab:scaling}
\begin{tabular}{@{}rrrrrl@{}}
\toprule
$C$ & mean $K$ & ratio & held-out CE & MATH-500 & Suite avg@8 [95\% CI] \\
\midrule
512 & 128 & 4:1 & 0.4227 & 88.3 & 41.5 {\scriptsize[37.6, 45.3]} \\
512 & 64 & 8:1 & 0.4281 & 86.7 & 39.4 {\scriptsize[35.5, 43.2]} \\
512 & 256 & 2:1 & 0.4184 & 87.6 & 43.4 {\scriptsize[39.4, 47.2]} \\
256 & 64 & 4:1 & 0.4246 & 86.9 & 40.7 {\scriptsize[36.7, 44.5]} \\
1024 & 256 & 4:1 & 0.4260 & 87.2 & 42.5 {\scriptsize[38.6, 46.3]} \\
\bottomrule
\end{tabular}
\end{table}

\label{app:blockcost}
\begin{figure}[h]
  \centering
  \includegraphics[width=0.55\linewidth]{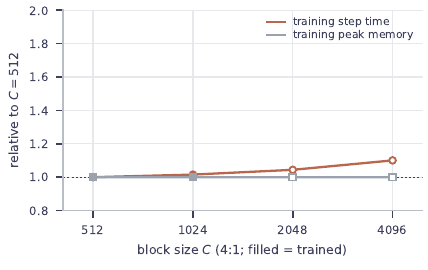}
  \caption{Training cost of the block size at 4:1, relative to $C{=}512$: step time and peak memory for one forward and
  backward pass over a 16K-token sequence on one H100 (filled: block sizes we trained). $C{=}256$, not shown, takes $1.9\times$
  as long, from partly filled attention tiles and twice as many blocks.}
  \label{fig:blockcost}
\end{figure}
Larger blocks cost little more to train (Figure~\ref{fig:blockcost}). With activation checkpointing, peak memory is set by the
sequence length and is the same at every block size; step time rises by 2\% at $C{=}1024$ and 10\% at $C{=}4096$, from the
attention inside each open block.

\FloatBarrier
\subsection{Coding domain at prototyping scale}
\label{app:code}
Every number in the main text is on mathematical reasoning. To test whether the recipe carries to another long-generation
domain we repeated the comparison on code at a reduced scale: 600M response tokens per run (0.3 of the main budget), with
held-out cross-entropy as the metric. The data are 120K problem--solution pairs from
OpenCodeReasoning~\citep{ahmad2025opencodereasoning}, each solution a DeepSeek-R1 reasoning trace of 512--16K tokens ending in a
Python program (median 7.0K tokens). At the same scale on mathematics, two uniform models that differ only in an initially
inactive projection differ by $0.0005$, which sets the resolution of the comparison.
\begin{table}[h]
\centering\small
\caption{Coding domain at prototyping scale: held-out cross-entropy (nats per response token) after 600M response tokens on OpenCodeReasoning, and the difference to the full-KV model trained identically. All compressed arms keep 128 code rows per 512-token block (4:1). Resolution of the comparison: 0.0005 (a null pair at the same scale on mathematics).}
\label{tab:code}
\begin{tabular}{lcc}
\toprule
Arm & Held-out CE & vs.\ full KV \\
\midrule
Full KV & 0.5450 & -- \\
\rowcolor{oursbg} Uniform, 128 rows & 0.5558 & $+0.0108$ \\
\rowcolor{oursbg} Per-layer, schedule measured on mathematics & 0.5544 & $+0.0094$ \\
\rowcolor{oursbg} Per-layer, schedule measured on code & 0.5544 & $+0.0094$ \\
\bottomrule
\end{tabular}
\end{table}

Table~\ref{tab:code} shows the same ordering as on mathematics: full KV, then per-layer, then uniform. Compression at 4:1
costs $0.011$ nats on code (2\% of the full-KV cross-entropy), and the per-layer schedule recovers $0.0014$ of it. Re-measuring
the schedule on the coding model gives the same layers to within one at each budget level (five of the six 256-row layers and
eleven of the twelve 64-row layers), and the two schedules reach the same held-out cross-entropy: the value a layer places on
its code rows is a property of the model rather than of the training domain. Accuracy and length on code remain to be measured,
which needs full-budget training and an execution-based benchmark.

\FloatBarrier
\section{Serving and correctness}
\label{app:serving}\label{app:efficiency}
\paragraph{Implementation.} The patch adds compressed decoding to vLLM's v1 engine. When a request completes a block of $C$
tokens, the scheduler appends its $K$ pause tokens; their KV at every layer is written into the block's persistent slots, and
the block's $C$ raw rows are freed. The pause tokens run in the same forward pass as other requests' decode tokens, which removes
a separate engine step; merged steps are captured as piecewise CUDA graphs, with the same capture range for every arm, baselines
included. A per-layer schedule is served as one KV cache group per quota level, so a layer with quota $q$ stores $q$ rows per
block and nothing else; uniform arms use a single group.

\paragraph{Time to finish.} Each arm's time to finish is taken at the best concurrency cap of a doubling grid. \method{} cannot
run 256 sequences on one H100 (at 128 its cache is already 92\% full), so its value is bounded by memory; the uniform setting,
which does run at 256, is slower there than at 128, so we expect the further gain to be small.

\paragraph{Parity.}\label{app:parity} Every accuracy in the paper is measured on the served model, so the serving path must
compute the same function as the trainer (Table~\ref{tab:parity}). The trainer's single forward pass under the block mask and an
incremental generator that physically deletes each completed block's KV agree on per-token cross-entropy to within
$4.5\times10^{-4}$; a deliberately wrong schedule differs by two orders of magnitude. On every serving path the vLLM engine
agrees with that generator on the top-1 token at 99.5\% or more of positions across block boundaries. As a negative control, we replay 6{,}000-token vLLM generations through
the same weights with full attention over the uncompressed history: on the tokens where that model and the compressing
generator disagree by more than 0.05 nats, the served model follows the compressing generator on 287 of 297 tokens (uniform),
469 of 479 (per-layer, masked) and 294 of 311 (per-layer, physical cache groups), so the served models do not attend to
compressed-away history.
\begin{table}[h]
\centering\small
\caption{Train/serve parity checks.}
\label{tab:parity}
\begin{tabular}{@{}p{0.30\linewidth} p{0.32\linewidth} p{0.31\linewidth}@{}}
\toprule
Check & Setting & Result \\
\midrule
Trainer vs.\ incremental generator (per-token CE) & vph checkpoint, 3 held-out sequences, trained schedule & mean $|\Delta|$ 3.5e-4 to 4.5e-4, max 1.3e-2 to 2.3e-2 \\
Same, deliberately wrong (uniform) schedule & control & mean $|\Delta|$ 1.6e-2 to 2.2e-2, max 4.0 \\
HF reference vs.\ vLLM packed KV (greedy tokens) & uniform128, hetero\_proxy, vph; 510 prompt + 16 generated (one block boundary) & identical greedy tokens \\
Multi-boundary HF vs.\ vLLM parity & per-layer headline, physical cache groups; 510-token prompt, 1,200 teacher-forced tokens (two block boundaries) & top-1 agreement 0.995, mean $|\Delta\log p|$ 0.006 \\
\bottomrule
\end{tabular}
\end{table}

\end{document}